\documentclass[journal,transmag]{IEEEtran}
\usepackage{cite}
\usepackage{caption}
\usepackage{graphicx}
\usepackage{algorithm}
\usepackage{algorithmic}
\usepackage{multirow}
\usepackage{amsmath,amssymb,amsfonts}
\usepackage{mathrsfs}
\usepackage{subfigure}
\newcommand{\tabincell}[2]{\begin{tabular}{@{}#1@{}}#2\end{tabular}}
\usepackage{booktabs}

\ifCLASSINFOpdf
\else
\fi
\begin{document}
	%
	\title{Generalized Label Enhancement with Sample Correlations}

	
	\author{\IEEEauthorblockN{Qinghai Zheng\IEEEauthorrefmark{1},
			Jihua Zhu\IEEEauthorrefmark{1}, Haoyu Tang\IEEEauthorrefmark{1}, Xinyuan Liu\IEEEauthorrefmark{1}, Zhongyu Li\IEEEauthorrefmark{1}, Huimin Lu\IEEEauthorrefmark{2}}
		\IEEEauthorblockA{\IEEEauthorrefmark{1}Lab of Vision Computing and Machine Learning, School of Software Engineering, \\Xi'an Jiaotong University, Xi’an 710049, China}
		\IEEEauthorblockA{\IEEEauthorrefmark{2}Environment Recognition \& Intelligent Computation Laboratory, Kyushu Institute of Technology, Japan}
		\thanks{Corresponding author: Jihua Zhu (email: zhujh@xjtu.edu.cn).}}

	\markboth{Journal of \LaTeX\ Class Files,~Vol.~12, No.~8, August~2015}%
	{Shell \MakeLowercase{\textit{et al.}}: Bare Demo of IEEEtran.cls for IEEE Transactions on Magnetics Journals}
	%



	\IEEEtitleabstractindextext{%
		\begin{abstract}
			Recently, label distribution learning (LDL) has drawn much attention in machine learning, where LDL model is learned from labelel instances. Different from single-label and multi-label annotations, label distributions describe the instance by multiple labels with different intensities and accommodate to more general scenes. Since most existing machine learning datasets merely provide logical labels, label distributions are unavailable in many real-world applications. To handle this problem, we propose two novel label enhancement methods, i.e., Label Enhancement with Sample Correlations (LESC) and generalized Label Enhancement with Sample Correlations (gLESC). More specifically, LESC employs a low-rank representation of samples in the feature space, and gLESC leverages a tensor multi-rank minimization to further investigate the sample correlations in both the feature space and label space. 
			Benefitting from the sample correlations, the proposed methods can boost the performance of label enhancement. Extensive experiments on 14 benchmark datasets demonstrate the effectiveness and superiority of our methods.
		\end{abstract}
		
		}

	\maketitle

	\IEEEdisplaynontitleabstractindextext

	%
	\IEEEpeerreviewmaketitle

	\section{Introduction}
	%
	%
	%
	%
	\IEEEPARstart{R}{ecently}, a growing number of studies have focused on the challenging label ambiguity learning problems. Since the single-label learning paradigm, in which an instance is mapped to one single logical label simply, has limitations in practice \cite{zhang2013review}, multi-label learning (MLL) is highlighted to address this issue. During past years, a collection of scenarios have applied this learning process \cite{tsoumakas2007multi,huang2012multi,chen2019multi}, which can simultaneously assign multiple logical labels to each instance. For example, in supervised MLL, each sample is described by a label vector, elements of which are either 1 or 0 to demonstrate whether this instance belongs to the corresponding label or not. Since multiple logical labels with the same values contribute equally for MLL, the relative importance among these multiple associated labels, which is supposed to be different under most circumstances, is ignored and cannot be well investigated.
	
	Therefore, despite MLL's success, in some sophisticated scenario, such as facial age estimation \cite{geng2013facial} and facial expression recognition \cite{jia2019facial}, the performance of primitive MLL is hindered, since a model precisely mapping the instance to a real-valued label vector with the quantitative description degrees, i.e., label distributions, is required in these tasks. To meet this demand, the learning process for the above-mentioned model termed "label distribution learning (LDL)"  \cite{geng2016label} has attracted significant attention. In LDL, an instance is annotated by the label vector, i.e., the label distribution, and each element ranging from 0 to 1 is the description degree of the relevant label and all values add up to 1. As many pieces of literature have demonstrated \cite{geng2016label,gao2017deep,zheng2018label}, label distributions can describe attributes of samples more precisely, since the relative importance of multiple labels is much different in many real-world applications, and implicit cues within the label distributions can be effectively used through LDL for reinforcing the supervised training.
	
	\begin{figure}[!htbp]
		\centering	
		\resizebox{.95\columnwidth}!{
			\includegraphics{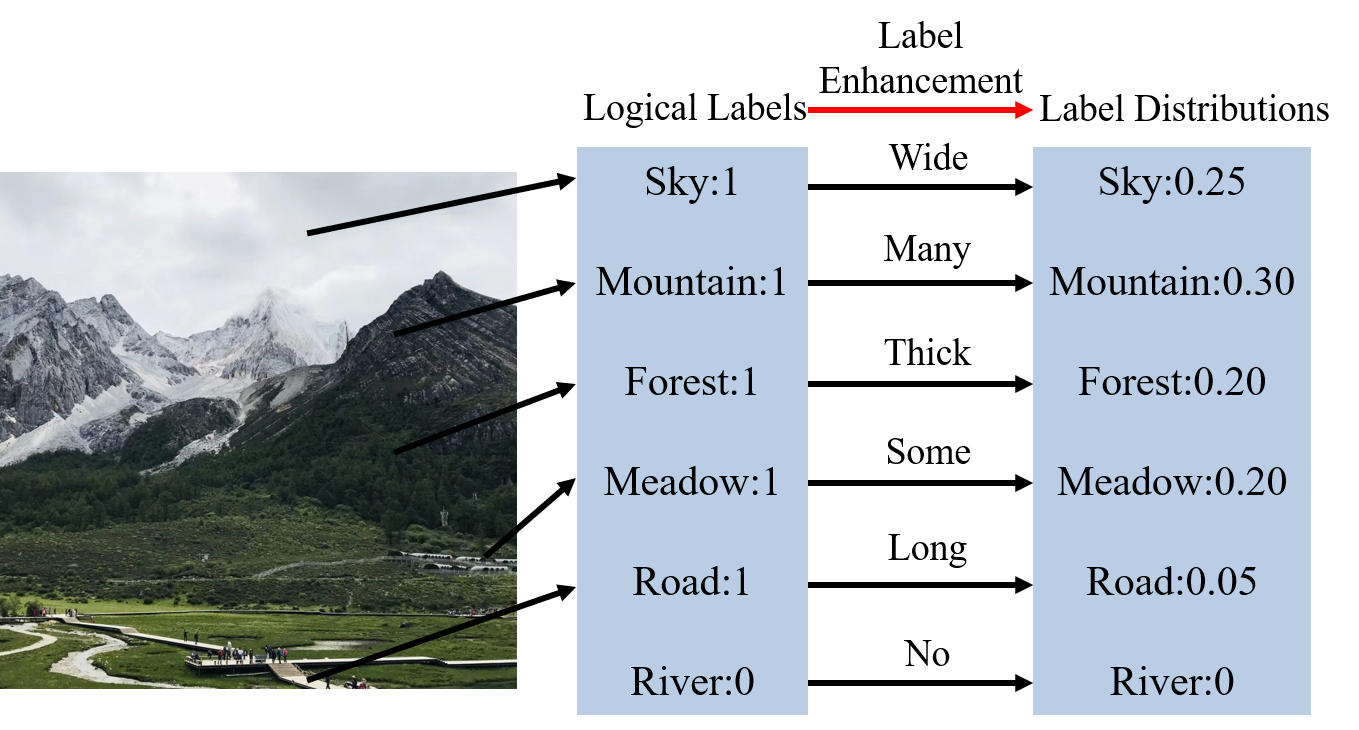}
		}
		\caption{An example of label enhancement}
		\label{Intro}
	\end{figure}
	
	
	Nevertheless, it is difficult to get samples with label distributions in practice. For example, SJAFFE dataset contains 213 grayscale images of 10 Japanese female models, each image is scored by 60 persons on the 6 basic emotions (i.e., happiness, sadness, surprise, fear, anger and disgust) with a five-level scale (1 represents the lowest emotion intensity, while 5 represents the highest emotion intensity). The average score (after normalization) of each emotion label is used to represent the label distribution \cite{geng2016label,xu2019partial}. Obviously, the manually annotating a dataset with label distributions is labor-intensive and time-consuming. Requirement of label distributions among different datasets drives progress in the label enhancement (LE), which is proposed by \cite{xu2018label,xu2019label}. More specifically, LE can be a pre-processing of LDL, in other words, label distributions can be exactly recovered from the off-the-shelf logical labels and the implicit information of the given features by LE, as shown in Fig.~\ref{Intro}.
	
	Obviously, according to the definition \cite{xu2018label,xu2019label}, the essence of recovering process is to utilize the information from two aspects: 1) the underlying topological structure in the feature space, and 2) the existing logical labels. Accordingly, several approaches have been proposed in recent years. However, most existing LE methods doesn't fully investigate and utilize both the underlying structures in the feature space and the implicit information of the existing logical labels. For example, graphs and similarity matrices utilized in \cite{li2015leveraging,hou2016multi,xu2018label} can not fully explore the intrinsic information of data samples, since edges in the graph or element of the similarity matrix are calculated by a pair-wise method \cite{li2015leveraging} or a $K$-nearest neighbors (KNN) strategy \cite{hou2016multi,xu2018label}. The downside of these partial-based graph construction processes is that only local topological features can be leveraged, and holistic information of the feature space is largely untapped. 
	
	Here, we aim to employ the intrinsic global sample correlations to obtain the exact recovery of label distributions. To get the global sample correlations, a choice is to construct the high-order similarity matrix, which assumes that samples with shared neighbors are likely to be similar. However, similar to the construction of similarity matrix, the construction of high-order similarity matrix require some prior knowledge for the graph construction, that is to say, if the parameter K of KNN is tuned slightly, the recovery performance of these algorithms may vary on a large scale, which is not expected in practice.  Since the low-rank representation (LRR) \cite{liu2012robust} can unearth the global structure of the whole feature space, it is expected to achieve a promising LE performance by employing LRR to supervise the label distribution recovering process. 
	
	Accordingly, a Label Enhancement with Sample Correlations, termed LESC, is proposed in this paper. To be specific, the proposed method imposes a low-rank constraint on the data subspace representation to capture the global relationship of all instances. Clearly, LRR is employed to benefit the LE performance by exploiting the intrinsic sample correlations in the feature space from a global perspective \cite{liu2011latent,yin2015laplacian,zheng2019feature,zheng2020constrained}. Since both labels are also the semantic features of data samples, it is natural and intuitive to transfer the constructed low-rank structure in the feature space to the label space smoothly. More importantly, by extending on the investigation of sample correlations employed in our previous work \cite{tang2020label}, this paper also proposes a generalized Label Enhancement with Sample Correlations, dubbed gLESC for short. This method can jointly investigate the implicit information in both the feature space and the label space by leveraging the tensor-Singular Value Decomposition (t-SVD) \cite{kilmer2013tsvd} based low-rank tensor constraint. Actually, the sample correlations simply obtained from the feature space is not the optimal choice for label distributions recovering, since excessive ineffective information, which is useless for LE, is also contained in the feature space. For example, regarding to the facial emotion labels, the sample correlations information of gender and identity in the feature space may hinder the recovering process of LE. To address this problem, the existing logical labels are also leverage to attain the desired and intrinsic sample correlations, which can be more suitable for LE. It is clear that samples with similar label distributions have similar logical labels, but not vice versa. Figuratively speaking, by imposing a t-SVD based low-rank tensor constraint on both the feature space and label space jointly, logical labels play a role to remove unwanted information. Once the desired sample correlations are attained, they are leveraged to supervise the recovering process of LE, and optimal recovered label distributions can be achieved after LE.
	
	Our contributions can be summarized as follows:
	\begin{itemize}
		\item [1)] 
		Different from the most existing approaches, we propose a novel Label Enhancement with Sample Correlations (LESC), which investigates the underlying global sample correlations in the feature space via the low-rank representation to recover the label distributions. The proposed LESC levels up the state-of-the-arts, which only explore the local sample correlations.  
		\item [2)]
		In addition to fully explore the global sample correlations, we introduce the tensor Singular Value Decomposition (t-SVD) based tensor multi-rank minimization to further unearth the global sample correlations in both the feature space and label space, and propose a generalized Label Enhancement with Sample Correlations (gLESC) in this paper. The strategy of leveraging the t-SVD based tensor multi-rank minimization fully mines the information of samples and make the recovery results more reliable. 
		\item [3)]
		Extensive experiments are conducted on 14 benchmark datasets, and experimental results validate the effectiveness and the superiority of our methods compared to several state-of-the-arts.
	\end{itemize}
	
	The remainder of this paper is organized as follows. The next section reviews related works of LE. Section 3 elaborates our proposed approaches, including LESC and gLESC. Comprehensive experimental results and corresponding discussions are provided in Section 4 and Section 5. Finally, conclusions of this paper are drawn in Section 6.

	\section{Related Work}
	\label{Section2}
	For the convenience of the description of related works, we declare the fundamental notations in advance. The set of labels is $Y=\left\{ {{y}_{1}},{{y}_{2}},\cdot \cdot \cdot ,{{y}_{o}} \right\}$, where $o$ is the size of the label set. For an instance ${{x}_{i}}\in {{\mathbb{R}}^{q}}$, the logical label is denoted as ${{L}_{i}}={{\left(l_{{{x}_{i}}}^{{{y}_{1}}},l_{{{x}_{i}}}^{{{y}_{2}}},\cdot \cdot \cdot ,l_{{{x}_{i}}}^{{{y}_{o}}}\right)}^{T}}$ and $l_{{{x}_{i}}}^{y}\in \text{ }\!\!\{\!\!\text{ 0,1 }\!\!\}\!\!\text{ }$, while the corresponding label distribution is denoted as:
	\begin{equation}
	{{D}_{i}}={{\left( d_{{{x}_{i}}}^{{{y}_{1}}},d_{{{x}_{i}}}^{{{y}_{2}}},\cdot \cdot \cdot ,d_{{{x}_{i}}}^{{{y}_{o}}} \right)}^{T}}, s.t., \sum\limits_{m=1}^{o}{d_{{{x}_{i}}}^{{{y}_{m}}}}\text{=}1,
	\end{equation}
	where $d_{{{x}_{i}}}^{y}$ depicts the degree to which ${{x}_{i}}$ belongs to label $y$. The goal of LE process is to recover the associated label distributions of every instance from the existing logical labels in a given dataset. This definition is formally presented by \cite{xu2019label}, in which a LE method, termed GLLE, is also proposed. It is worth noting that some studies have concentrated on the same issue earlier. We categorize existing related approaches into algorithm adaptation, which extends certain existing methods naturally to deal with the problems of label enhancement, and specialized algorithms, which are specially designed for label enhancement. 
	
	\subsection{Algorithm Adaptation}
	Some existing methods are extended to tackle the problems of label enhancement straightforward, such as fuzzy clustering method (FCM) \cite{melin2005hybrid,el2006study}, which employs the fuzzy C-means clustering and intends to allocate the description values to each instance over diverse clusters, and kernel method (KM), which  leverages the fuzzy SVM to learn the membership degrees of data samples.  
	
	For FCM, it uses the fuzzy C-means clustering to get different clusters and cluster prototypes, then obtains membership degree of each data sample with respect to different cluster prototypes, finally labels all samples with label distributions by employing the fuzzy composition and softmax normalization. Specifically, samples in the feature space are clustered into $t$ clusters via the fuzzy C-means clustering where ${{c}_{k}}$ denotes the $k$-th cluster center. Different from the k-means clustering, the fuzzy C-means clustering computes coefficients of being in the clusters for each data point, and we denote the coefficients as cluster memberships or membership degrees here. For for each instance ${{x}_{i}}$, the membership degrees ${{\omega }_{i}}=\left\{ {{\omega }_{i1}},{{\omega }_{i2}},\cdot \cdot \cdot ,{{\omega }_{it}} \right\}$ with respect to ${{c}_{k}}$ and can be calculated as follows:
	\begin{equation}
	{{\omega }_{ik}}=\frac{1}{\sum\limits_{j=1}^{t}{{{\left( \frac{{{\left\| {{x}_{i}}-{{c}_{k}} \right\|}_{2}}}{{{\left\| {{x}_{i}}-{{c}_{j}} \right\|}_{2}}} \right)}^{\frac{1}{\beta -1}}}}},
	\end{equation}
	where $\beta$ is larger than 1 and is the hyper-parameter used to controls how fuzzy the cluster will be. The
	higher it is, the fuzzier the cluster will be in the end. Then to get the soft connections between classes
	and clusters, a prototype label matrix is constructed by 
	\begin{equation}
	\label{FCM_Q}
	{{Q}_{j}}={{Q}_{j}}+{{\omega }_{i}},s.t.,~l_{{{x}_{i}}}^{{{y}_{j}}}=1, 
	\end{equation}
	where ${{Q}_{j}}$ denotes the $j$-th row of $Q$. For $Q\in {{\mathbb{R}}^{o\times t}}$, it is initialized as a zero matrix, and can be continuously updated according to Eq.~(\ref{FCM_Q}). After normalizing the columns and rows of $Q$ to sum to 1, the label distribution is computed for each instance ${{x}_{i}}$ using fuzzy composition: ${{D}_{i}}=Q\circ {{\omega }_{i}}$.
	
	Regarding to KM, it utilizes the fuzzy membership function in a fuzzy SVM \cite{KM}. For a specific logical label, taking $l_{x_i}^{y_m}$ for example, the dataset $S=\{({{x}_{1}},{{L}_{1}}),({{x}_{2}},{{L}_{2}}),\cdot \cdot \cdot ,({{x}_{n}},{{L}_{n}})\}$ is separated into two parts,  including the $C_{+}^{y_m}$ and $C_{-}^{y_m}$. It uses $p_{+}^{y_m}$ to denote the corresponding center of $C_{+}^{y_m}$:
	\begin{equation}
	p_ + ^{{y_m}}{\text{ = }}\frac{1}{{{n_ + }}}\sum\limits_{x \in C_ + ^{{y_m}}} {f({x_i})},  \hfill
	\end{equation}
	where $n_+$ indicates the sample numbers in the $C_{+}^{y_m}$, and $f( \cdot )$ denotes a nonlinear function. The corresponding radius and the distance between $x_i$ and $p_{+}^{y_m}$ can be also obtained as follows:
	\begin{equation}
	\begin{gathered}
	{r_ + } = \max \left\| {p_ + ^{{y_m}} - f({x_i})} \right\|, \hfill \\
	dis{t_i} = \left\| {f({x_i}) - p_ + ^{{y_m}}} \right\|. \hfill \\ 
	\end{gathered}
	\end{equation}
	Consequently,  the corresponding label distribution of a sample, e.g., $x_i$, to $l_{x_i}^{y_m}$ label, e.g., $l_{x_i}^{y_m}$, can be achieved by:
	\begin{equation}
	d_{{x_i}}^{{y_m}} = \left\{ \begin{gathered}
	1 - (\frac{{dist_i^2}}{{r_ + ^2 + \eta }}),{\rm{if}}{\kern 1pt} {\kern 1pt} {\kern 1pt} l_{{x_i}}^{{y_m}} = 1, \hfill \\
	0,{\rm{if}}{\kern 1pt} {\kern 1pt} {\kern 1pt} l_{{x_i}}^{{y_m}} = 0, \hfill \\ 
	\end{gathered}  \right.
	\end{equation}
	where $\eta$ is larger than 0, and the desired label distributions can be obtained with a softmax normalization to let $\sum\nolimits_{m=1}^{o}{d_{{{x}_{i}}}^{{{y}_{m}}}}\text{=}1$.
	
	Although the topological structures of the feature space can be explored by fuzzy C-means clustering in FCM and fuzzy SVM in KM, they lack of a good investigation of the sample correlations, and the information in both the feature and label space can not be fully explored as well.
	
	\subsection{Specialized Algorithms}
	Recently, some specialized algorithms \cite{li2015leveraging,hou2016multi,xu2018label} are also proposed, and they obtain label enhancement by using the graph information during the learning process. For example, the label propagation (LP) technique, which is common used in semisupervised learning, is utilized to get label enhancement \cite{li2015leveraging}. Based on manifold learning, ML  learns the topological structures of the feature
	space by the KNN linear combination firstly, and then constrains the process of label enhancement \cite{hou2016multi}. According to the assumption that two instances, which are close to each other in the feature space, are more likely to share the same label, the graph Laplacian label enhancement (GLLE) utilizes the graph information in the feature space to improve the recovery process of label distributions \cite{xu2018label}.
	
	For LP, the label propagation process \cite{zhu2009introduction} is used to achieve label enhancement. The pairwise similarity is calculated over the complete feature space, and then a fully-connected graph is established as follows:
	\begin{equation}
	{{q}_{ij}}=\left\{ \begin{aligned}
	& \exp \left( -\frac{{{\left\| {{x}_{i}}-{{x}_{j}} \right\|}^{2}}}{2{{\sigma }^{2}}} \right),{\rm{if}}~i\ne j, \\ 
	& 0,{\rm{if}}~i=j, \\ 
	\end{aligned} \right.
	\end{equation}
	where $\forall ~i,j\in \![1,n]\!$ and $\sigma$ is fixed be 1. For the required LP matrix, it can be built from the formula: $P\!=\!{{\tilde{Q}}^{-\frac{1}{2}}}Q{{\tilde{Q}}^{-\frac{1}{2}}}$ with $\tilde{Q}\!=\!\rm{diag}\left[ {{{\tilde{q}}}_{1}},{{{\tilde{q}}}_{2}},\cdot \cdot \cdot ,{{{\tilde{q}}}_{n}} \right]$ denoting a diagonal matrix where ${{\tilde{q}}_{i}}$ equals to the sum of $i$-th row element in $Q$. Thus far, The LP is iteratively implemented, and it is proved that the recovered label distribution matrix $\mathfrak{D}=\left[ {{D}_{1}};{{D}_{2}};\cdot \cdot \cdot ;{{D}_{n}} \right]$ converges to:
	\begin{equation}
	{{\mathfrak{D}}^{*}}=\left( 1-\alpha  \right){{\left( I-\alpha P \right)}^{-1}}\Gamma,
	\end{equation}
	with $\alpha$ denoting the trade-off parameter to control the contribution between the label propagation $P$ and the initial logical label matrix $\Gamma$.
	
	Regarding to ML, it recovers the label distribution based on the manifold learning, which ensures them to gradually convert the local structure of the feature space into the label space. In particular, to represent this structure, the similarity matrix $Q$ is established based on the assumption that each feature can be represented by the linear combination of its KNN, which means to minimize: 
	\begin{equation}
	\Phi (Q)=\sum\limits_{i=1}^{n}{{{\left\| {{x}_{i}}-\sum\limits_{j\ne i}{{{q}_{ij}}}{{x}_{j}} \right\|}^{2}}},
	\end{equation}
	where ${{q}_{ij}}=1$, if ${{x}_{j}}$ belongs to the KNNs of ${{x}_{i}}$; otherwise, ${{q}_{ij}}=0$. Then they further
	constrain that $\sum_{j=1}^{n} q_{i j}=1$ for translation invariance. The constructed graph is transferred into the label space to minimize the distance between the target label distribution and the identical linear combination of its KNN label distributions \cite{roweis2000nonlinear}, which infers the optimization of:
	\begin{equation}
	\phi (D)=\sum\limits_{i=1}^{n}{{{\left\| {{D}_{i}}-\sum\limits_{j\ne i}{{{q}_{ij}}}{{D}_{j}} \right\|}^{2}}}
	\end{equation}
	by adding the constraint of  $\forall 1\le i\le n,1\le j\le o,d_{xi}^{j}l_{i}^{j}\ge \lambda $, where $\lambda >0$. This formula is minimized with respect to the target label distribution $D$ through a constrained quadratic programming process.
	
	For the GLLE algorithm, the similarity matrix is also constructed in the feature space by partial topological structure. Different from LP, which calculates the pair-wise distance within the whole feature space, the GLLE algorithm computes the distance between a specific instance and its KNNs to define the relevant element in the similarity matrix as follows:
	\begin{equation}
	{{q}_{ij}}=\left\{ \begin{aligned}
	& \exp \left( -\frac{{{\left\| {{x}_{i}}-{{x}_{j}} \right\|}^{2}}}{2{{\sigma }^{2}}} \right),{\rm{if}}~{x}_{j}\in K\left( i \right), \\ 
	& 0,~{\rm{otherwise}}, \\ 
	\end{aligned} \right.
	\end{equation}
	where  $K\left( i \right)$ is the set of ${x}_{i}$'s KNNs. Then GLLE incorporates  the constructed graph into the label space to learn a mapping model, which can get the desired label distributions effectively.
	
	It can be observed that these methods, including LP, ML, and GLLE, utilize the graph information in the feature space to guide the learning process of label enhancement. However, only local information is investigated in these methods, and the information hidden in the logical labels is also not fully investigated and leveraged for label enhancement.
	
	
	Since intrinsic sample relationships may not be investigated by local graph information fully, the similarity matrix based on the pair-wise or local feature structure hinders label recovery performance. Here, the LRR and the t-SVD based low-rank tensor constraint are introduced to excavate the global information and to leverage the attained proper sample correlations, so as to overcome these aforementioned drawbacks in the label distribution recovering process.
	
	\section{Our Proposed Approaches}
	In this section, our methods, i.e., LESC and gLESC, are introduced detailed. In a training set $S=\{({{x}_{1}},{{L}_{1}}),({{x}_{2}},{{L}_{2}}),\cdot \cdot \cdot ,({{x}_{n}},{{L}_{n}})\}$, all instances are vertically concatenated along the column to attain the feature matrix $X=[{{x}_{1}};{{x}_{2}};\cdot \cdot \cdot {{x}_{n}}]$, where ${{x}_{i}}\in {{\mathbb{R}}^{q}}$ and $X\in {{\mathbb{R}}^{q\times n}}$. After the LE process, a new LDL training set $\varepsilon =\{({{x}_{1}},{{D}_{1}}),({{x}_{2}},{{D}_{2}}),\cdot \cdot \cdot ,({{x}_{n}},{{D}_{n}})\}$ can be rehabilitated to implement the LDL process. Here we use $\Gamma =\left[ {{L}_{1}};{{L}_{2}};\cdot \cdot \cdot ;{{L}_{n}} \right]$ and $\mathfrak{D}=\left[ {{D}_{1}};{{D}_{2}};\cdot \cdot \cdot ;{{D}_{n}} \right]$ to denote the logical label matrix and the objective label distribution matrix respectively.
	
	\subsection{LESC Approach}
	
	\begin{figure}[!t]
		\centering	
		\resizebox{.95\columnwidth}!{
			\includegraphics{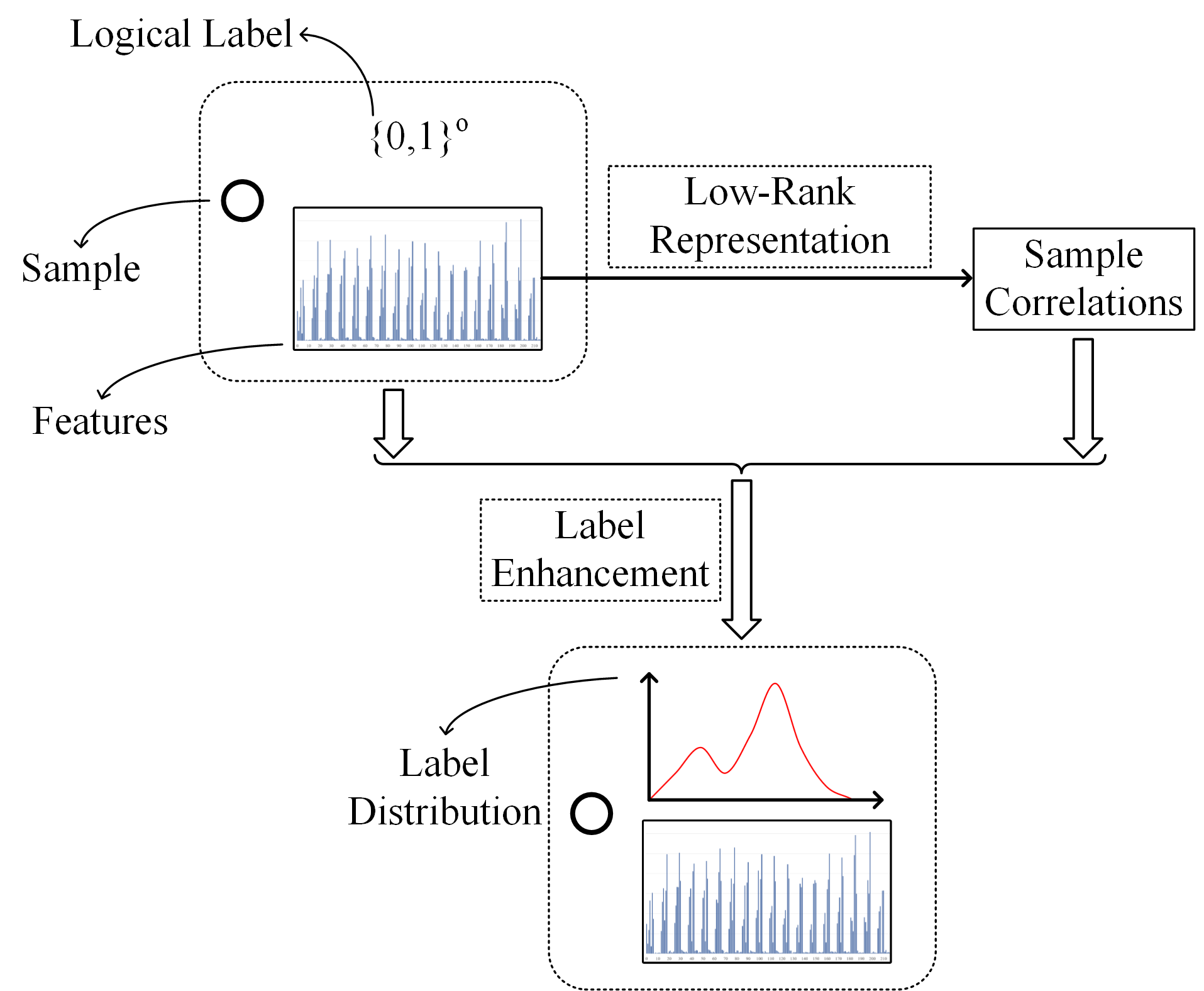}
		}
		\caption{The flow chart of the proposed LESC.}
		\label{LESC}
	\end{figure}
	
	For a given instance ${{x}_{i}}$, it is necessary to find an effective model to recover the desired label distribution. The mapping model employed in this paper can be written as follows:
	\begin{equation}
	\label{eq.10}
	{{D}_{i}}=\phi \left( \hat{\theta },\xi \left( {{x}_{i}} \right) \right),
	\end{equation}
	where $\phi (\hat{\theta },\cdot  )$ indicates the linear transformation parameterized by $\hat \theta  = [\theta ,b] = [{\theta ^1},{\theta ^2}, \cdots ,{\theta ^o},b]$:
	\begin{equation}
	\phi \left( {\hat \theta ,\xi \left( {{x_i}} \right)} \right) = {\theta ^T}\xi \left( {{x_i}} \right) + b,
	\end{equation}
	where $\theta$ is a weight matrix, $b$ is a bias vector, and $\xi \left( x \right)$ embeds $x$ into a high-dimensional space, in which the Gaussian kernel function is determined to be employed.
	
	To get the optimal $\hat{\theta }$, the following objective function can be formulated:
	\begin{equation}
	\label{objective func}
	\underset{{\hat{\theta }}}{\mathop{\min }}\,\mathcal{L}\left( {\hat{\theta }} \right)+{\lambda}_{1} \Psi \left( {\hat{\theta }} \right),
	\end{equation}
	where $\mathcal{L}( {\hat{\theta }} )$ denotes a loss function, $\Psi ( {\hat{\theta }} )$ is used to excavate the underlying information of sample correlations, and ${\lambda}_{1} $ is a trade-off parameter. To be specific, we will elaborate $\mathcal{L}( {\hat{\theta }} )$ and $\Psi ( {\hat{\theta }} )$ detailed in this section.
	
	For $\mathcal{L}( {\hat{\theta }} )$, it is the loss function between the recovered label distributions and the logical labels. The behind reason is that a recovered label distributions should be close to existing logical labels. Taking a sample with the logical label $\{0,1,1\}$ and the recovered label distribution $\{d^1,d^2,d^3\}$ for example, it is very reasonable to assume that $d^1$ is close to $0$, $d^2$ and $d^3$ are both near to $1$, since the prior knowledge of the ground-truth label distribution is unavailable. Therefore, $\mathcal{L}( {\hat{\theta }} )$ can be formulated as follows:
	\begin{equation}
	\label{FirstTerm}
	\mathcal{L}\left( {\hat{\theta }} \right)=\sum\limits_{i=1}^{n}{{{\left\| \phi \left( \hat{\theta },\xi \left( {{x}_{i}} \right) \right)-{{L}_{i}} \right\|}^{2}}},
	\end{equation}
	in which the least-squares (LS) loss function is adopted here.
	
	Regarding to $\Psi ( {\hat{\theta }} )$, it plays an important role in the recovery process of label distributions. To be specific, the global sample correlations are employed:
	\begin{equation}
	\label{SecondTerm}
	\Psi \left( {\hat{\theta }} \right)=\left\| {\mathfrak{D}}-{\mathfrak{D}}{\hat{C}} \right\|_{F}^{2} \text{=}\left\| ( I-{{ {\hat{C}} }^{T}} )\mathfrak{D}^{T} \right\|_{F}^{2}, 
	\end{equation}
	where $\hat{C}$ is the minimized LRR of the feature space. Different from existing methods, which employ the local graph structure (e.g. KNN) during the recovery process, the low-rank representation (LRR) is applied
	in $\Psi ( {\hat{\theta }} )$ to get global topological structure in the feature space to guide the recovery process. 
	Since data samples can be expressed by the linear combination of related samples, the global sample correlations in the feature space can be explored by LRR. The motivation under Eq.~(\ref{SecondTerm}) is that leveraging the global topological information in feature space can improve the performance of LE. Therefore, the low-rank representation of feature space plays a critical role as the prior information to benefit the process of LE. Therefore, it is expected that the low-rank recovery to the label distribution ${\mathfrak{D}}$ can be expressed, which means to discover a proper ${\mathfrak{D}}$ for minimizing the distance between ${\mathfrak{D}}$ and ${\mathfrak{D}}{\hat{C}}$. 
	
	To be clear, the flow chart of LESC is presented in Fig.~\ref{LESC}. As can be observed, the sample correlations are obtained by applying the low-rank representation on the feature space. In other words, the proposed LESC aims at seeking the LRR of the feature matrix to excavate the global structure in the feature space. Consequently, by assuming that $X=XC+E$, it is natural and necessary to solve the following rank minimization problem:
	\begin{equation}
	\underset{C,E}{\mathop{\min }}\,{\rm{rank}}\left( C \right)+{\lambda}_{2} {{\left\| E \right\|}_{l}}, s.t., X=XC+E,
	\label{LRR_LESC}
	\end{equation}
	where $E$ indicates the sample-specific corruptions, and ${\lambda}_{2}$ is the low-rank coefficients which balances the effects between two parts. $\hat{C}$ is used to denote the desired low-rank representation of feature \textit{X} with respect to the variable $C$. Practically, the rank function can be replaced by a nuclear norm to transfer (\ref{LRR_LESC}) into a convex optimization problem. As a result, we have the following problem:
	\begin{equation}
	\underset{C,E}{\mathop{\min }}\,{{\left\| C \right\|}_{*}}+{\lambda}_{2} {{\left\| E \right\|}_{2,1}}, s.t., X=XC+E,
	\label{LESC_objective}
	\end{equation}
	where  ${\left\| C \right\|_ * }$ denotes the nuclear norm of matrix $C$. Specifically, it can be defined as follows:
	\begin{equation}
	{\left\| C \right\|_ * } = \sum\limits_{i = 1}^{n} {{\sigma _i}},
	\end{equation}
	where ${\sigma _i}$ is the $i$th singular value of the matrix $C$. The nuclear norm ${\left\| C \right\|_ * }$ is a convex envelope of the rank function $\rm{rank}{(C)}$, so it can be used in mathematical optimization to search for the low rank matrix. Since corruptions of a dataset happen on a small fraction of data samples, so we term such corruptions, i.e., $E$, as “sample-specific corruptions”, and the $l_{2,1}$-norm is used to handle this kind corruption:
	\begin{equation}
	{\left\| E \right\|_{2,1}} = \sum\limits_{j = 1}^n {\sqrt {\sum\limits_{i = 1}^q {E_{ij}^2} } },
	\end{equation}
	which is the sum of $l_2$ norm of all columns of the matrix $E$.
	
	To get optimal solution, the augmented Lagrange multiplier (ALM) with Alternating Direction Minimization strategy \cite{lin2011ALM} is employed in this paper. Specifically, an auxiliary variable, i.e., $J$, is introduced here so as to make the objective function separable and convenient for optimization. Therefore, (\ref{LESC_objective}) can be rewritten as follows:
	\begin{equation}
	\begin{matrix}
	\underset{J,C,E}{\mathop{\min }}\,{{\left\| J \right\|}_{*}}+{\lambda}_{2} {{\left\| E \right\|}_{2,1}} \\ 
	s.t.~X=XC+E, C=J,\\ 
	\end{matrix}
	\end{equation}
	and the corresponding ALM problem can be solved by minimizing the following function:
	\begin{equation}
	\label{ALMLESC}
	\begin{aligned}
	&\rm{ALM}(J,C,E,{Y_1},{Y_2},\mu )=\|J{{\|}_{*}}+{\lambda}_{2} \|E{{\|}_{2,1}} \\ 
	& \!+\!\left\langle {{Y_1},X - X{C} - {E}} \right\rangle \!+\!\left\langle {{Y_2},C-J} \right\rangle \\ 
	& \!+\!\frac{\mu }{2}\left( \|X-XC-E\|_{F}^{2}\!+\!\|C-J\|_{F}^{2} \right), 
	\end{aligned}
	\end{equation}
	which can be solved by an off-the-shelve method, i.e., the augmented Lagrange multiplier (ALM) with Alternating Direction Minimization. We skip this optimization process over for the compactness of this paper. 
	
	Once Eq.~(\ref{ALMLESC}) is optimized, the desired sample correlations can be utilized for label enhancement. Consequently, Eq.~(\ref{objective func}) can be rewritten as follows:
	\begin{equation}
	\begin{aligned}
	& \underset{{\hat{\theta }}}{\mathop{\min }}\sum\limits_{i=1}^{n}{{{\left\| \phi \left( \hat{\theta },\xi_{i} \right)\!-\!{{L}_{i}} \right\|}^{2}}}\!+\!{\lambda}_{1}\left\| \left( I\!-\!{{\hat{C}}^{T}} \right){\mathfrak{D}^{T}} \right\|_{F}^{2}\\ 
	&~~~\!=\!{\rm{tr}}\left[ {{\left( \phi \left( \hat{\theta },\Xi  \right)\!-\!\Gamma  \right)}^{T}}\left( \phi \left( \hat{\theta },\Xi  \right)\!-\!\Gamma  \right) \right]\\
	&~~~~~~ +{\lambda}_{1} {\rm{tr}}\left(\mathfrak{D}\left( I\!-\!\hat{C} \right)\left( I\!-\!{\hat{C}^{T}} \right){{\mathfrak{D}}^{T}} \right),  
	\end{aligned}
	\label{get_distribution}
	\end{equation}
	where $\Xi =\left[ \xi \left( {{x}_{1}} \right),\cdot \cdot \cdot ,\xi \left( {{x}_{n}} \right) \right]$. Aiming to achieve the optimal solution, i.e., ${{\hat{\theta }}^{*}}$, the optimization of Eq.~(\ref{get_distribution}) can be solved by an effective quasi-Newton method called the limited memory BFGS (L-BFGS) \cite{nocedal2006numerical}, of which the optimizing process can be associated with the first-order gradient. Similar to the solution of LRR, it is an off-the-shelve method, and we skip it over as well.
	
	Once the formula converges, we feed the optimal ${{\hat{\theta }}^{*}}$ into Eq.~(\ref{eq.10}) to form the label distribution ${{D}_{i}}$. Furthermore, since the defined label distribution needs to meet the requirement $\sum\nolimits_{m=1}^{o}{d_{{{x}_{i}}}^{{{y}_{m}}}}\text{=}1$, ${{D}_{i}}$ is normalized by the softmax normalization form.
	
	
	\subsection{gLESC Approach}
	
	\begin{figure}[!t]
		\centering	
		\resizebox{.95\columnwidth}!{
			\includegraphics{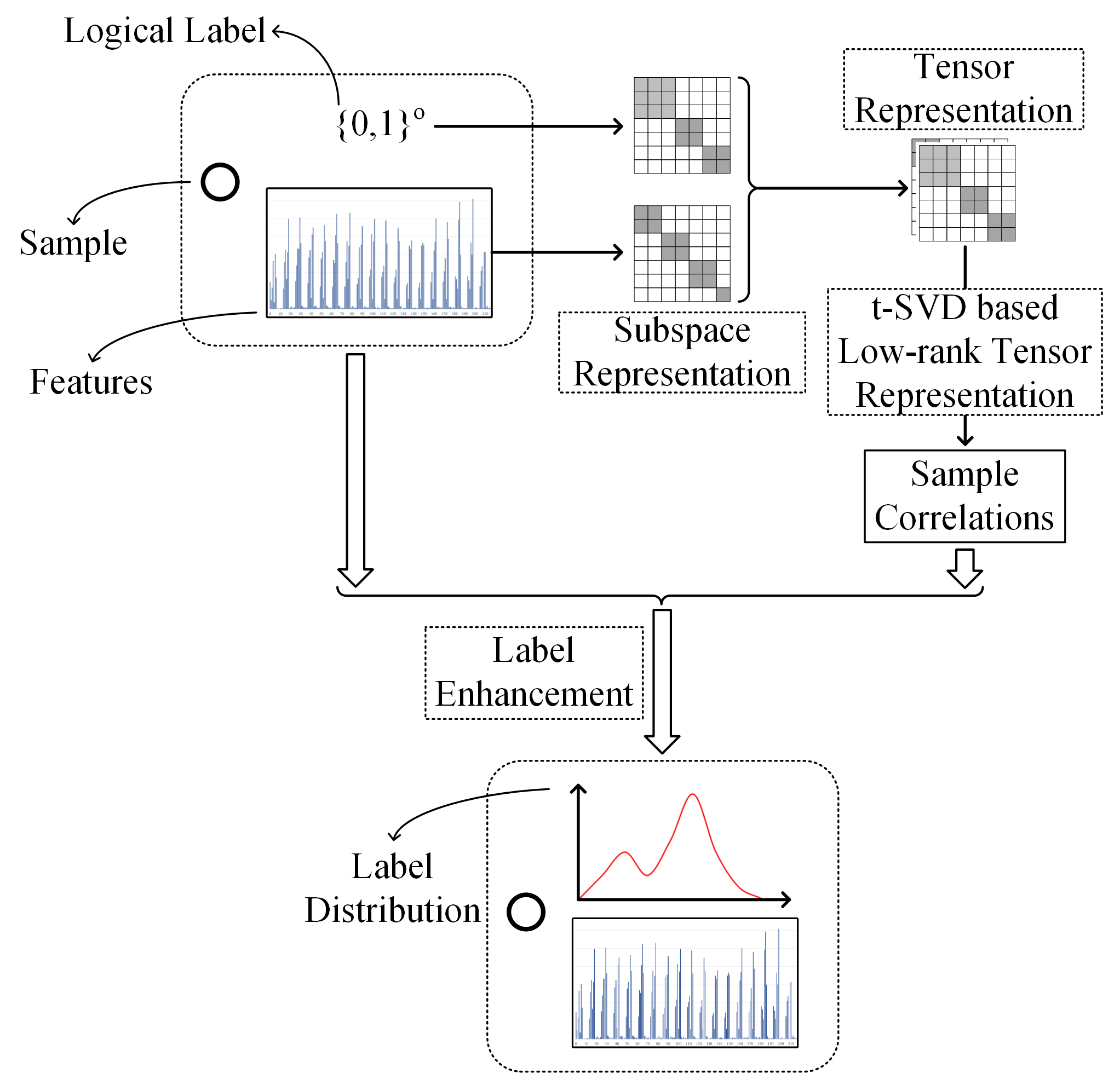}
		}
		\caption{The flow chart of the proposed gLESC.}
		\label{gLESC}
	\end{figure}
	
	LESC employs a low-rank subspace representation in the feature space to get the sample correlations, which can be utilized to supervise the recovering process of label distributions. Under the assumption that both the features and the labels are all the semantic information of samples, it is natural and reasonable to impose the aforementioned constraint on the desired label distributions. However, only the sample correlations in the feature space are investigated in the proposed LESC, and the corresponding information hidden in the existing logical labels is ignored. Actually, the sample correlations obtained by LRR in the feature space are influenced by some interference information. For example, in the facial emotion dataset, the gender and identity information may be also contained in the sample correlations, which are attained by LRR in the feature space. Obviously, it obstructs the exact recovering process of label distributions. 
	
	Since the existing logical labels do not contain the unwished information, it is a good choice to incorporate the underlying information of these existing logical labels into the formation process of the desired sample correlations. To this end, a generalized label enhancement with sample correlations (gLESC) is also proposed in this paper, and the corresponding flow chart is shown in Fig.~\ref{gLESC}. As can be observed in Fig.~\ref{gLESC}, the underlying sample correlations of both the sample features and existing logical label can be achieved to supervise the whole recovering process of label distributions, so the refinement of LE can be attained, as well as the implicit information of data samples can be leveraged fully. To achieve this goal, the 
	tensor-Singular Value Decomposition (t-SVD) based low-rank tensor constraint \cite{kilmer2013tsvd} is introduced in this section. It should be noted that the difference between LESC and gLESC is the construction of sample correlations. 
	
	To be clear, we emphatically introduce how to use the t-SVD based low-rank tensor constraint to incorporate the underlying information of logical labels into formation process of sample correlations. Specifically, in the proposed gLESC, we have the following formulation:
	\begin{equation}
	\begin{aligned}
	&\mathop {\min }\limits_{{\mathcal{C}},{\mathcal{E}}} {\left\| {\mathcal{C}} \right\|_ \circledast } + {\lambda _2}{\left\| {\mathcal{E}} \right\|_{2,1}},\\
	&s.t.~X = X{C^{(1)}} + {E^{(1)}},\\
	&~~~~~\Gamma  = \Gamma {C^{(2)}} + {E^{(2)}},
	\end{aligned}
	\label{tensor_obj}
	\end{equation}
	where ${\mathcal{C}}$ and ${\mathcal{E}}$ are 3-order tensors constructed by ${\left\{ {{C^{(i)}}} \right\}_{i = 1,2}}$ and ${\left\{ {{E^{(i)}}} \right\}_{i = 1,2}}$, respectively. ${\left\| {\mathcal{C}} \right\|_ \circledast }$ denotes a t-SVD based tensor nuclear norm \cite{kilmer2013tsvd}, which can be calculated as follows:
	\begin{equation}
	{\left\| {\mathcal{C}} \right\|_ \circledast } = \sum\limits_{i = 1}^2 {{{\left\| {{\mathcal{C}}_f^{(i)}} \right\|}_ * }}  = \sum\limits_{k = 1}^n {\sum\limits_{i = 1}^2 {\left| {\Sigma _f^{(i)}(k,k)} \right|} },
	\end{equation}
	in which ${{\mathcal{C}}_f^{(i)}}$ denotes a fast Fourier transformation (FFT) along the 3-rd dimension of ${\mathcal{C}}$, i.e, the frontal slices of ${\mathcal{C}}$ \cite{zhang2014novel}, and ${\Sigma _f^{(i)}(k,k)}$ indicates the \textit{k}-th diagonal element of ${\Sigma _f^{(i)}}$, which can be calculated as follows:
	\begin{equation}
	{\mathcal{C}}_f^{(i)} = U_f^{(i)}\Sigma _f^{(i)}V{_f^{(i)T}}.
	\end{equation}
	According to the unitary invariance of the matrix nuclear norm, we can reformulate ${\left\| {\mathcal{C}} \right\|_ \circledast }$ as follows:
	\begin{equation}
	{\left\| {\mathcal{C}} \right\|_ \circledast }={\left\| {\operatorname{bdiag} ({\Sigma _f})} \right\|_ * } = {\left\| {\operatorname{bdiag} ({\mathcal{C}_f})} \right\|_ * }
	\end{equation}
	where $\rm{bdiag}({\mathcal{C}_f})$ unfolds ${\mathcal{C}_f}$ to the matrix with the following block-diagonal form:
	\begin{equation}
	{\operatorname{bdiag}}({\mathcal{C}_f}){\text{ = }}\left[ {\begin{array}{*{20}{c}}
		{\mathcal{C}_f^{(1)}}&{} \\ 
		{}&{\mathcal{C}_f^{(2)}} 
		\end{array}} \right].
	\end{equation}
	Considering the Fourier transform, the block circulant matrix can be block diagonalized straightforward, and then, we have the following formulation:
	\begin{equation}
	{\left\| {{\operatorname{bdiag}}({\mathcal{C}_f})} \right\|_ * } = {\left\| {\operatorname{bcirc} (\mathcal{C})} \right\|_ * },
	\end{equation}
	where $\rm{bcirc} (\mathcal{C})$ of can be formulated as follows:
	\begin{equation}
	\operatorname{bicrc} (\mathcal{C}) = \left[ {\begin{array}{*{20}{c}}
		{{C^{(1)}}}&{{C^{(2)}}} \\ 
		{{C^{(2)}}}&{{C^{(1)}}} 
		\end{array}} \right].
	\end{equation}
	Consequently, minimizing ${\left\| {\mathcal{C}} \right\|_ \circledast }$ can be rewritten as follows:
	\begin{equation}
	\label{TSVDDecom}
	\mathop {\min }\limits_{\mathcal{C}} {\left\| {\left[ {\begin{array}{*{20}{c}}
				{{C^{(1)}}}&{{C^{(2)}}} \\ 
				{{C^{(2)}}}&{{C^{(1)}}} 
				\end{array}} \right]} \right\|_ * }.
	\end{equation}
	
	As can be observed in Eq.~(\ref{TSVDDecom}), by using the t-SVD based low-rank tensor constraint, we can incorporate the underlying information of the logical labels into the formation process of sample correlations, as well as the information in the feature space.
	
	To solve the minimization problem of (\ref{tensor_obj}), we can construct the following ALM equation:
	\begin{equation}
	\begin{aligned}
	&\rm{ALM}({\mathcal{G}}, {\mathcal{C}},{\mathcal{E}} )={\left\| {\mathcal{G}} \right\|_ \circledast } + {\lambda _2}{\left\| {\mathcal{E}} \right\|_{2,1}} \\ 
	& \!+\!\left\langle {{Y_1},X - X{C^{(1)}} - {E^{(1)}}} \right\rangle\!+\!\frac{\mu }{2}\left( \|X-XC^{(1)}-E^{(1)}\|_{F}^{2}\right)\\ 
	& \!+\!\left\langle {{Y_2},\Gamma - \Gamma{C^{(2)}} - {E^{(2)}}} \right\rangle \!+\!\frac{\mu }{2}\left(\|\Gamma-\Gamma{C}^{(2)}-E^{(2)}\|_{F}^{2}\right)\\ 
	& \!+\!\left\langle {{\mathcal{W}},{\mathcal{C}} - {\mathcal{G}}} \right\rangle  + \frac{\rho }{2}\left\| {{\mathcal{C}} - {\mathcal{G}}} \right\|_F^2, 
	\end{aligned}
	\end{equation}
	where ${\mathcal{G}}$ is an auxiliary tensor variable, $Y_1$, $Y_2$ and ${\mathcal{W}}$ are the Lagrange multipliers. Since the corresponding optimization algorithm is an off-the-shelve method, we also skip it over.
	
	Once the problem of Eq.~(\ref{tensor_obj}) is optimized, we can obtain desired sample correlations as follows so as to achieve an exact recovering process of label distributions:
	\begin{equation}
	\hat C = \frac{1}{2}({C^{(1)}} + {C^{(2)}}).
	\end{equation}
	
	The subsequent operation of gLESC is similar to Eq.~ (\ref{get_distribution}). For the compactness of this paper, we skip them over in this section.
	
	\begin{figure}[!t]
		\centering	
		\resizebox{1\columnwidth}!{
			\includegraphics{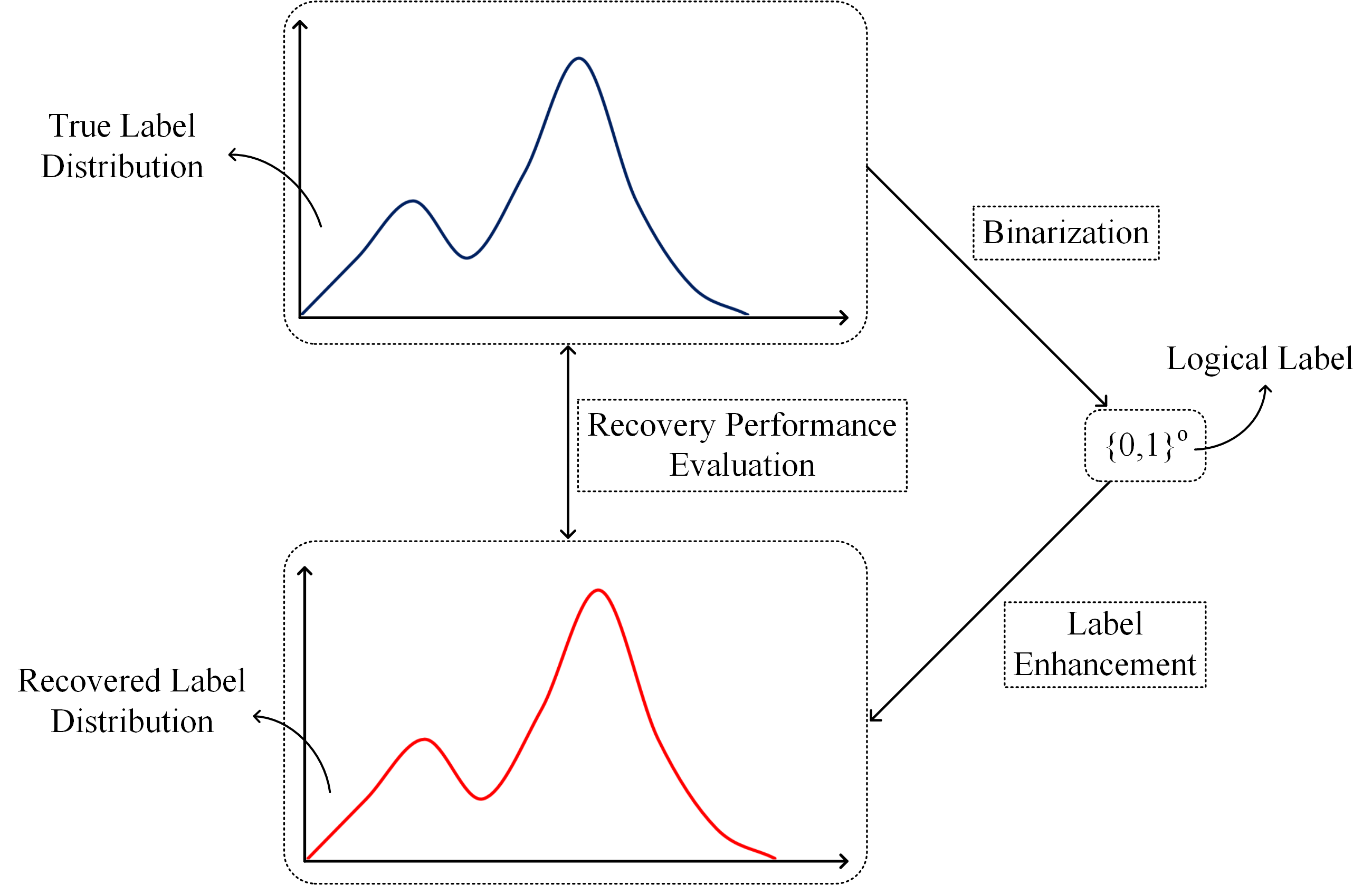}
		}
		\caption{The flowchart of label recovery experiments conducted in this section. To be specific, the true label distributions are binarized to attain the logical labels firstly, then a LE method can be performed to obtain the recovered label distributions. Accordingly, we evaluate the recovered performance based on six frequently-used LDL evaluation measures \cite{geng2016label}.}
		\label{Label_Recovery_Flowchart}
	\end{figure}
	
	\section{Experiments}
	To validate the effectiveness and superiority of the proposed LESC and gLESC, extensive experiments are conducted, and the corresponding experimental results are also presented in this section. To be specific, the label recovery experiments are performed on 14 benchmark datasets$\footnote{http://palm.seu.edu.cn/xgeng/LDL/index.htm}$, and the corresponding flowchart is shown in Fig.~\ref{Label_Recovery_Flowchart}. 
	
	\begin{figure*}[!t]
		\centering	
		\resizebox{1\textwidth}!{
			\includegraphics{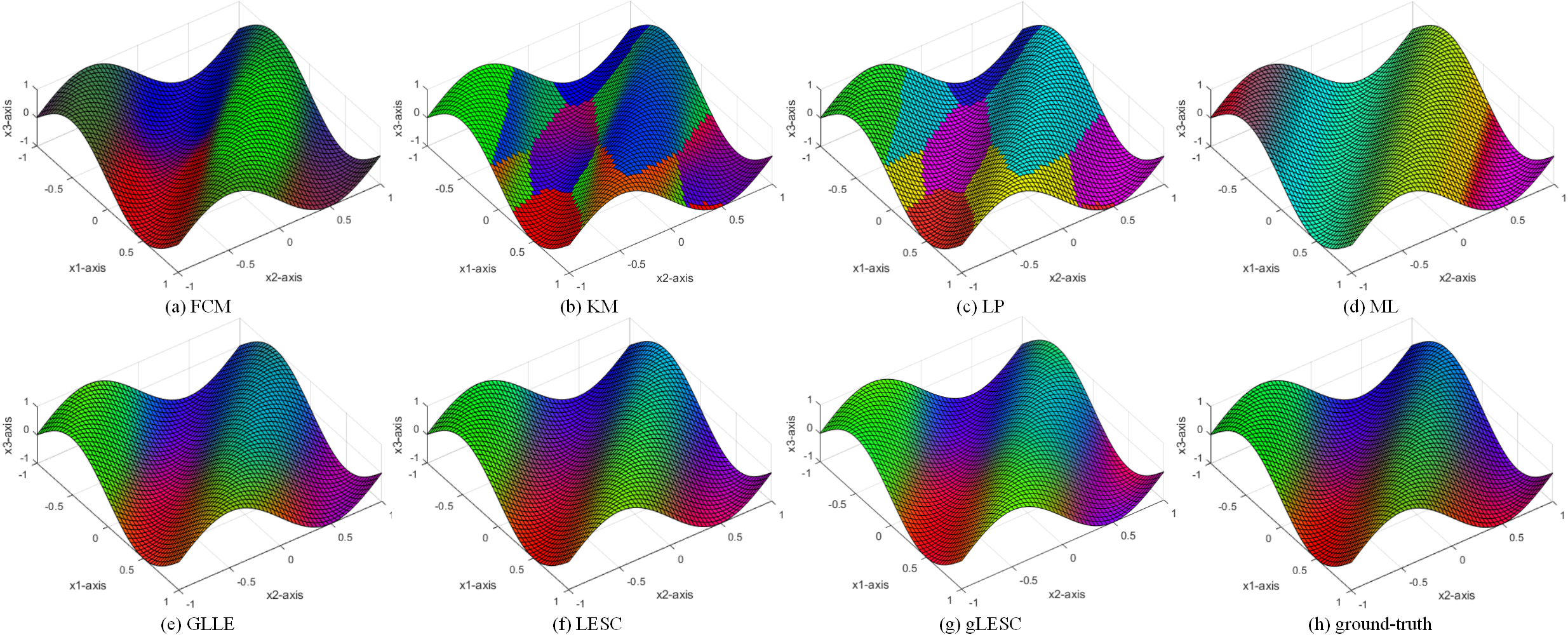}
		}
		\caption{Visualization of the ground-truth and recovered label distributions on the artificial dataset (regarded as RGB colors, best viewed in color).}
		\label{Art_3D}
	\end{figure*}
	
	\begin{table}	
		\centering
		\caption{Some Information about 14 Datasets.}	
		\begin{center}
			\resizebox{.9 \columnwidth}!{
				\begin{tabular}{c|c|c|c}
					\toprule
					Dataset &  \#~Instances & \#~Features & \#~Labels\\
					\midrule
					\tabincell{c}{Artificial\\Movie\\SBU\_3DFE\\SJAFFE\\Yeast-alpha\\Yeast-cdc\\Yeast-cold\\Yeast-diau\\Yeast-dtt\\Yeast-elu\\Yeast-heat\\Yeast-spo\\Yeast-spo5\\Yeast-spoem} &  \tabincell{c}{2601\\7755\\2500\\213\\2465\\2465\\2465\\2465\\2465\\2465\\2465\\2465\\2465\\2465} & \tabincell{c}{3\\1869\\243\\243\\24\\24\\24\\24\\24\\24\\24\\24\\24\\24} &
					\tabincell{c}{3\\5\\6\\6\\18\\15\\4\\7\\4\\14\\6\\6\\3\\2} \\
					\bottomrule
			\end{tabular}}
		\end{center}
		\label{dataset_statistics}		
	\end{table}
	
	\begin{table}
		\centering
		\caption{Introduction to Evaluation Measures.}
		\label{Evaluation_Measures}	
		\begin{center}
			\resizebox{.9\columnwidth}!{
				\begin{tabular}{c|c}
					\toprule
					Measure &  Formula \\
					\midrule
					Cheb $\downarrow$ &  $Di{{s}_{1}}(D,\hat{D})={{\max }_{j}}\left| {{d}^{{{y}_{j}}}}-{{{\hat{d}}}^{{{y}_{j}}}} \right|$ \\
					\midrule
					Canber $\downarrow$ & $Di{{s}_{2}}(D,\hat{D})=\sum\limits_{j=1}^{o}{\frac{\left| {{d}^{{{y}_{j}}}}-{{{\hat{d}}}^{{{y}_{j}}}} \right|}{{{d}^{{{y}_{j}}}}+{{{\hat{d}}}^{{{y}_{j}}}}}}$ \\
					\midrule
					Clark $\downarrow$ & $Di{{s}_{3}}(D,\hat{D})=\sqrt{\sum\limits_{j=1}^{o}{\frac{{{\left( {{d}^{{{y}_{j}}}}-{{{\hat{d}}}^{{{y}_{j}}}} \right)}^{2}}}{{{\left( {{d}^{{{y}_{j}}}}+{{{\hat{d}}}^{{{y}_{j}}}} \right)}^{2}}}}}$ \\
					\midrule
					KL $\downarrow$ & 
					$Di{s_4}(D,\widehat D) = \sum\limits_{j = 1}^o {{d^{{y_j}}}\ln \frac{{{d^{{y_j}}}}}{{{{\widehat d}^{{y_j}}}}}} $ \\ 
					\midrule
					Cosine $\uparrow$ &  $Si{{m}_{1}}(D,\hat{D})=\frac{\sum\limits_{j=1}^{o}{{{d}^{{{y}_{j}}}}}{{{\hat{d}}}^{{{y}_{j}}}}}{\sqrt{\sum\limits_{j=1}^{o}{{{\left( {{d}^{{{y}_{j}}}} \right)}^{2}}}}\sqrt{\sum\limits_{j=1}^{o}{{{\left( {{{\hat{d}}}^{{{y}_{j}}}} \right)}^{2}}}}}$ \\
					\midrule
					Intersec $\uparrow$ & $Si{{m}_{2}}(D,\hat{D})=\sum\limits_{j=1}^{o}{\min }\left( {{d}^{{{y}_{j}}}},{{{\hat{d}}}^{{{y}_{j}}}}\right)$ \\	
					\bottomrule
			\end{tabular}}
		\end{center}	
	\end{table}
	
	\subsection{Datasets}
	\begin{table*}
		\centering
		\caption{Recovery Results (value(rank)).}
		\label{Recovery_Results_Cheb_Canber}
		\begin{center}
			\resizebox{1\textwidth}!{
				\begin{tabular}{c|c|c|c|c|c|c|c||c|c|c|c|c|c|c}
					\toprule
					\multirow{2}{*}{Dataset} &
					\multicolumn{7}{c||}{Measure Results by Cheb $\downarrow $}&
					\multicolumn{7}{c}{Measure Results by Canber $\downarrow $}\\
					\cmidrule{2-15}
					& FCM & KM & LP & ML & GLLE & LESC & gLESC & FCM & KM & LP & ML & GLLE & LESC & gLESC\\
					\midrule
					\tabincell{c}{Artificial\\Movie\\SBU\_3DFE\\SJAFFE\\Yeast-alpha\\Yeast-cdc\\Yeast-cold\\Yeast-diau\\Yeast-dtt\\Yeast-elu\\Yeast-heat\\Yeast-spo\\Yeast-spo5\\Yeast-spoem\\} &  		\tabincell{c}{0.188(5)\\0.230(6)\\0.135(5)\\0.132(5)\\0.044(5)\\0.051(5)\\0.141(5)\\0.124(5)\\0.097(4)\\0.052(5)\\0.169(6)\\0.130(5)\\0.162(5)\\0.233(5)\\}&  \tabincell{c}{0.260(7)\\0.234(7)\\0.238(7)\\0.214(7)\\0.063(7)\\0.076(7)\\0.252(7)\\0.152(7)\\0.257(7)\\0.078(7)\\0.175(7)\\0.175(7)\\0.277(7)\\0.408(7)\\} &
					\tabincell{c}{0.130(4)\\0.161(4)\\0.123(2)\\0.107(4)\\0.040(4)\\0.042(4)\\0.137(4)\\0.099(4)\\0.128(5)\\0.044(4)\\0.086(4)\\0.090(4)\\0.114(4)\\0.163(4)\\} &
					\tabincell{c}{0.227(6)\\0.164(5)\\0.233(6)\\0.186(6)\\0.057(6)\\0.071(6)\\0.242(6)\\0.148(6)\\0.244(6)\\0.072(6)\\0.165(5)\\0.171(6)\\0.273(6)\\0.403(6)\\} &
					\tabincell{c}{0.108(3)\\0.122(3)\\0.126(4)\\0.087(3)\\0.020(3)\\0.022(3)\\0.066(3)\\0.053(3)\\0.052(3)\\0.023(3)\\0.049(3)\\0.062(3)\\0.099(3)\\0.088(3)\\}&
					\tabincell{c}{0.057(2)\\0.121(2)\\\bf{0.122(1)}\\0.069(2)\\0.015(2)\\0.019(2)\\0.056(2)\\0.042(2)\\0.043(2)\\0.019(2)\\0.046(2)\\0.060(2)\\\bf{0.092(1)}\\0.087(2)\\} & \tabincell{c}{\bf{0.055(1)}\\\bf{0.120(1)}\\0.125(3)\\\bf{0.067(1)}\\\bf{0.014(1)}\\\bf{0.017(1)}\\\bf{0.052(1)}\\\bf{0.039(1)}\\\bf{0.037(1)}\\\bf{0.017(1)}\\\bf{0.043(1)}\\\bf{0.059(1)}\\\bf{0.092(1)}\\\bf{0.084(1)}\\} &
					\tabincell{c}{0.797(5)\\1.664(4)\\1.020(4)\\1.081(5)\\2.883(4)\\2.415(4)\\0.734(4)\\1.895(4)\\0.501(4)\\1.689(4)\\1.157(4)\\0.998(4)\\0.563(5)\\0.534(5)\\} &
					\tabincell{c}{1.779(7)\\3.444(7)\\4.121(7)\\4.010(7)\\11.809(7)\\9.875(7)\\2.566(7)\\4.261(7)\\2.594(7)\\9.110(7)\\3.849(7)\\3.854(7)\\1.382(7)\\1.253(7)\\} &
					\tabincell{c}{0.668(4)\\1.720(5)\\1.245(5)\\1.064(4)\\4.544(5)\\3.644(5)\\0.924(5)\\1.748(5)\\0.941(5)\\3.381(5)\\1.293(5)\\1.231(5)\\0.401(4)\\0.365(4)\\} &
					\tabincell{c}{1.413(6)\\1.934(6)\\4.001(6)\\3.138(6)\\11.603(6)\\9.695(6)\\2.519(6)\\4.180(6)\\2.549(6)\\8.949(6)\\3.779(6)\\3.772(6)\\1.355(6)\\1.226(6)\\} & 
					\tabincell{c}{0.617(3)\\1.045(3)\\0.820(3)\\0.781(3)\\1.134(3)\\0.959(3)\\0.305(3)\\0.671(3)\\0.248(3)\\0.902(3)\\0.430(3)\\0.548(3)\\0.305(3)\\0.183(3)\\} & 
					\tabincell{c}{0.213(2)\\\bf{1.034(1)}\\\bf{0.799(1)}\\0.561(2)\\0.846(2)\\0.765(2)\\0.263(2)\\0.480(2)\\0.206(2)\\0.727(2)\\0.401(2)\\0.533(2)\\0.284(2)\\0.180(2)\\} & 
					\tabincell{c}{\bf{0.193(1)}\\\bf{1.034(1)}\\0.803(2)\\\bf{0.550(1)}\\\bf{0.761(1)}\\\bf{0.695(1)}\\\bf{0.242(1)}\\\bf{0.452(1)}\\\bf{0.175(1)}\\\bf{0.628(1)}\\\bf{0.372(1)}\\\bf{0.521(1)}\\\bf{0.283(1)}\\\bf{0.175(1)}\\}
					\\
					\midrule
					Avg.Rank & 5.07 & 7.00 & 3.93 & 5.86 & 3.07 & 1.86 & \bf{1.14} & 4.27 & 7.00 & 4.71 & 6.00 & 3.00 & 1.86 & \bf{1.07}\\
					\midrule
					\multirow{2}{*}{Dataset} &
					\multicolumn{7}{c||}{Measure Results by Clark $\downarrow $}&
					\multicolumn{7}{c}{Measure Results by KL $\downarrow $}\\
					\cmidrule{2-15}
					& FCM & KM & LP & ML & GLLE & LESC & gLESC & FCM & KM & LP & ML & GLLE & LESC & gLESC\\
					\midrule
					\tabincell{c}{Artificial\\Movie\\SBU\_3DFE\\SJAFFE\\Yeast-alpha\\Yeast-cdc\\Yeast-cold\\Yeast-diau\\Yeast-dtt\\Yeast-elu\\Yeast-heat\\Yeast-spo\\Yeast-spo5\\Yeast-spoem\\} &  \tabincell{c}{0.561(5)\\0.859(4)\\0.482(4)\\0.522(5)\\0.821(4)\\0.739(4)\\0.433(4)\\0.838(5)\\0.329(4)\\0.579(4)\\0.580(5)\\0.520(4)\\0.395(5)\\0.401(5)\\}&  \tabincell{c}{1.251(7)\\1.766(7)\\1.907(7)\\1.874(7)\\3.153(7)\\2.885(7)\\1.472(7)\\1.886(7)\\1.477(7)\\2.768(7)\\1.802(7)\\1.811(7)\\1.059(7)\\1.028(7)\\} &
					\tabincell{c}{0.487(4)\\0.913(5)\\0.580(5)\\0.502(4)\\1.185(5)\\1.014(5)\\0.503(5)\\0.788(4)\\0.499(5)\\0.973(5)\\0.568(4)\\0.558(5)\\0.274(4)\\0.272(4)\\} &
					\tabincell{c}{1.041(6)\\1.140(6)\\1.848(6)\\1.519(6)\\3.088(6)\\2.825(6)\\1.440(6)\\1.844(6)\\1.446(6)\\2.711(6)\\1.764(6)\\1.768(6)\\1.036(6)\\1.004(6)\\} &
					\tabincell{c}{0.452(3)\\0.569(3)\\0.391(3)\\0.377(3)\\0.337(3)\\0.306(3)\\0.176(3)\\0.296(3)\\0.143(3)\\0.295(3)\\0.213(3)\\0.266(3)\\0.197(3)\\0.132(3)\\}&
					\tabincell{c}{0.148(2)\\0.564(2)\\0.378(2)\\0.276(2)\\0.253(2)\\0.251(2)\\0.152(2)\\0.224(2)\\0.119(2)\\0.241(2)\\0.199(2)\\0.258(2)\\0.185(2)\\0.129(2)\\} & \tabincell{c}{\bf{0.130(1)}\\\bf{0.563(1)}\\\bf{0.376(1)}\\\bf{0.270(1)}\\\bf{0.231(1)}\\\bf{0.231(1)}\\\bf{0.141(1)}\\\bf{0.211(1)}\\\bf{0.102(1)}\\\bf{0.213(1)}\\\bf{0.186(1)}\\\bf{0.253(1)}\\\bf{0.184(1)}\\\bf{0.126(1)}\\} &
					\tabincell{c}{0.267(5)\\0.381(6)\\0.094(3)\\0.107(5)\\0.100(4)\\0.091(4)\\0.113(5)\\0.159(5)\\0.065(4)\\0.059(4)\\0.147(5)\\0.110(5)\\0.123(5)\\0.208(5)\\} &
					\tabincell{c}{0.309(7)\\0.452(7)\\0.603(6)\\0.558(7)\\0.630(7)\\0.630(7)\\0.586(7)\\0.538(7)\\0.617(7)\\0.617(7)\\0.586(7)\\0.562(7)\\0.334(7)\\0.531(7)\\} &
					\tabincell{c}{0.160(4)\\0.177(5)\\0.105(4)\\0.077(4)\\0.121(5)\\0.111(5)\\0.103(4)\\0.127(4)\\0.103(5)\\0.109(5)\\0.089(4)\\0.084(4)\\0.042(4)\\0.067(4)\\} &
					\tabincell{c}{0.274(6)\\0.218(4)\\0.565(5)\\0.391(6)\\0.602(6)\\0.601(6)\\0.556(6)\\0.509(6)\\0.586(6)\\0.589(6)\\0.556(6)\\0.532(6)\\0.317(6)\\0.503(6)\\} & 
					\tabincell{c}{0.131(3)\\0.123(3)\\0.069(2)\\0.050(3)\\0.013(3)\\0.014(3)\\0.019(3)\\0.027(3)\\0.013(3)\\0.013(3)\\0.017(3)\\0.029(3)\\0.034(3)\\0.027(2)\\} & 
					\tabincell{c}{0.013(2)\\\bf{0.120(1)}\\0.064(1)\\0.029(2)\\0.008(2)\\0.010(2)\\0.015(2)\\0.017(2)\\0.010(2)\\0.009(2)\\0.015(2)\\0.028(2)\\\bf{0.031(1)}\\0.027(2)\\} & 
					\tabincell{c}{\bf{0.012(1)}\\\bf{0.120(1)}\\\bf{0.064(1)}\\\bf{0.027(1)}\\\bf{0.007(1)}\\\bf{0.008(1)}\\\bf{0.013(1)}\\\bf{0.015(1)}\\\bf{0.007(1)}\\\bf{0.007(1)}\\\bf{0.014(1)}\\\bf{0.027(1)}\\\bf{0.031(1)}\\\bf{0.026(1)}\\}
					\\
					\midrule
					Avg.Rank & 4.43 & 7.00 & 4.57 & 6.00 & 3.00 & 2.00 & \bf{1.00} & 4.64 & 7.00 & 4.36 & 5.79 & 2.86 & 1.86 & \bf{1.00}\\
					\midrule
					\multirow{2}{*}{Dataset} &
					\multicolumn{7}{c||}{Measure Results by Cosine $\uparrow $}&
					\multicolumn{7}{c}{Measure Results by Intersec $\uparrow $}\\
					\cmidrule{2-15}
					& FCM & KM & LP & ML & GLLE & LESC & gLESC & FCM & KM & LP & ML & GLLE & LESC & gLESC\\
					\midrule
					\tabincell{c}{Artificial\\Movie\\SBU\_3DFE\\SJAFFE\\Yeast-alpha\\Yeast-cdc\\Yeast-cold\\Yeast-diau\\Yeast-dtt\\Yeast-elu\\Yeast-heat\\Yeast-spo\\Yeast-spo5\\Yeast-spoem\\} &  \tabincell{c}{0.933(5)\\0.773(7)\\0.912(5)\\0.906(5)\\0.922(4)\\0.929(4)\\0.922(5)\\0.882(5)\\0.959(4)\\0.950(4)\\0.883(5)\\0.909(5)\\0.922(5)\\0.878(5)\\}&  \tabincell{c}{0.918(7)\\0.880(6)\\0.812(7)\\0.827(7)\\0.751(7)\\0.754(7)\\0.779(7)\\0.799(7)\\0.759(7)\\0.758(7)\\0.779(7)\\0.800(7)\\0.882(7)\\0.812(7)\\} &
					\tabincell{c}{0.974(4)\\0.929(4)\\0.922(4)\\0.941(4)\\0.911(5)\\0.916(5)\\0.925(4)\\0.915(4)\\0.921(5)\\0.918(5)\\0.932(4)\\0.939(4)\\0.969(4)\\0.950(4)\\} &
					\tabincell{c}{0.925(6)\\0.919(5)\\0.815(6)\\0.857(6)\\0.756(6)\\0.759(6)\\0.784(6)\\0.803(6)\\0.763(6)\\0.763(6)\\0.783(6)\\0.803(6)\\0.884(6)\\0.815(6)\\} &
					\tabincell{c}{0.980(3)\\0.936(3)\\0.927(3)\\0.958(3)\\0.987(3)\\0.987(3)\\0.982(3)\\0.975(3)\\0.988(3)\\0.987(3)\\0.984(3)\\0.974(3)\\0.971(3)\\0.978(2)\\}&
					\tabincell{c}{\bf{0.992(1)}\\0.937(2)\\\bf{0.932(1)}\\0.973(2)\\0.992(2)\\0.991(2)\\0.986(2)\\0.985(2)\\0.991(2)\\0.991(2)\\0.986(2)\\0.975(2)\\\bf{0.974(1)}\\0.978(2)\\} & \tabincell{c}{0.991(2)\\\bf{0.938(1)}\\\bf{0.931(2)}\\\bf{0.975(1)}\\\bf{0.994(1)}\\\bf{0.992(1)}\\\bf{0.988(1)}\\\bf{0.987(1)}\\\bf{0.994(1)}\\\bf{0.993(1)}\\\bf{0.987(1)}\\\bf{0.976(1)}\\\bf{0.974(1)}\\\bf{0.979(1)}\\} &
					\tabincell{c}{0.812(5)\\0.677(6)\\0.827(4)\\0.821(5)\\0.844(4)\\0.847(4)\\0.833(4)\\0.760(5)\\0.894(4)\\0.883(4)\\0.807(4)\\0.836(4)\\0.838(5)\\0.767(5)\\} &
					\tabincell{c}{0.740(7)\\0.649(7)\\0.579(7)\\0.593(7)\\0.532(7)\\0.533(7)\\0.559(7)\\0.588(7)\\0.541(7)\\0.539(7)\\0.559(7)\\0.575(7)\\0.724(7)\\0.592(7)\\} &
					\tabincell{c}{0.870(4)\\0.778(5)\\0.810(5)\\0.837(4)\\0.774(5)\\0.779(5)\\0.794(5)\\0.788(4)\\0.786(5)\\0.782(5)\\0.805(5)\\0.819(5)\\0.886(4)\\0.837(4)\\} &
					\tabincell{c}{0.773(6)\\0.779(4)\\0.587(6)\\0.661(6)\\0.537(6)\\0.538(6)\\0.565(6)\\0.593(6)\\0.546(6)\\0.544(6)\\0.564(6)\\0.580(6)\\0.727(6)\\0.597(6)\\} & 
					\tabincell{c}{0.892(3)\\0.831(3)\\0.850(3)\\0.872(3)\\0.938(3)\\0.937(3)\\0.924(3)\\0.906(3)\\0.939(3)\\0.936(3)\\0.929(3)\\0.909(3)\\0.901(3)\\0.912(3)\\} & 
					\tabincell{c}{0.943(2)\\\bf{0.833(1)}\\\bf{0.855(1)}\\0.905(2)\\0.953(2)\\0.950(2)\\0.935(2)\\0.933(2)\\0.949(2)\\0.949(2)\\0.934(2)\\0.912(2)\\\bf{0.908(1)}\\0.913(2)\\} & 
					\tabincell{c}{\bf{0.945(1)}\\\bf{0.833(1)}\\0.854(2)\\\bf{0.908(1)}\\\bf{0.958(1)}\\\bf{0.954(1)}\\\bf{0.940(1)}\\\bf{0.937(1)}\\\bf{0.957(1)}\\\bf{0.956(1)}\\\bf{0.939(1)}\\\bf{0.914(1)}\\\bf{0.908(1)}\\\bf{0.916(1)}\\}
					\\
					\midrule
					Avg.Rank & 4.86 & 6.93 & 4.29 & 5.93 & 2.93 & 1.79 & \bf{1.14} & 4.50 & 7.00 & 4.64 & 5.86 & 3.00 & 2.13 & \bf{1.07}\\
					\bottomrule
				\end{tabular}
			}
		\end{center}
	\end{table*}
	
	The fundamental statistics of 14 datasets, including 13 real-world datasets and one toy dataset, employed for evaluation can be observed in Table~\ref{dataset_statistics}. To be specific, the first 3 real-world datasets are created from movies, facial expression images, the remaining 10 real-world datasets from Yeast-alpha to Yeast-spoem are collected from the records of some biological experiments on the budding yeast genes\cite{eisen1998cluster}. As for the artificial dataset, which is also adopted in \cite{xu2018label} to intuitively exhibits the model's ability of label enhancement, each instance ${x}_{i}\in {{\mathbb{R}}^{3}}$ is chosen following the rule that the first two dimensions $x_{i}^{\left( 1 \right)}$ and $x_{i}^{\left( 2 \right)}$ are formed as a grid with an interval of 0.04 in the range [-1,1], while the third dimension $x_{i}^{\left( 3 \right)}$ is computed by:
	\begin{equation}
	x_{i}^{\left( 3 \right)}=\sin \left( \left( x_{i}^{\left( 1 \right)}+x_{i}^{\left( 2 \right)} \right)\times \pi  \right)
	\end{equation}
	The corresponding label distribution ${{D}_{i}}={{\left(d_{{{x}_{i}}}^{{{y}_{1}}},d_{{{x}_{i}}}^{{{y}_{2}}},d_{{{x}_{i}}}^{{{y}_{3}}} \right)}^{T}}$  is collected through the following equations:
	\begin{equation}
	{{w}_{j}}\!=\!mx_{i}^{\left( j \right)}+n{{\left( x_{i}^{\left( j \right)} \right)}^{2}}+p{{\left( x_{i}^{\left( j \right)} \right)}^{3}}+q,j\!=\!1,2,3
	\end{equation}
	\begin{equation}
	\left\{ \begin{aligned}
	& {{\varphi }_{1}}={{\left( \textbf{r}_{1}^{\top }\textbf{w} \right)}^{2}} \\ 
	& {{\varphi }_{2}}={{\left( \textbf{r}_{2}^{\top }\textbf{w}+{{\eta }_{1}}{{\varphi }_{1}} \right)}^{2}} \\ 
	& {{\varphi }_{3}}={{\left( \textbf{r}_{3}^{\top }\textbf{w}+{{\eta }_{2}}{{\varphi }_{2}} \right)}^{2}} \\ 
	\end{aligned} \right.
	\end{equation}
	and label distributions can be obtained as follows:
	\begin{equation}
	d_{{{x}_{i}}}^{{{y}_{j}}}=\frac{{{\varphi }_{j}}}{{{\varphi }_{1}}+{{\varphi }_{2}}+{{\varphi }_{3}}},j=1,2,3
	\end{equation}
	where $\textbf{w}=({{w}_{1}},{{w}_{2}},{{w}_{3}})$, $m=1$, $n=0.5$, $p=0.2$, $q=1$, ${\textbf{r}_{1}}={{\left( 4,2,1 \right)}^{T}}$, ${ \textbf{r}_{2}}={{\left( 1,2,4 \right)}^{T}}$, ${\textbf{r}_{3}}={{\left( 1,4,2 \right)}^{T}}$, and ${{\eta }_{1}}={{\eta }_{2}}=0.01$.
	
	It is noteworthy that due to the lack of datasets with both logical labels and label distributions, the logical labels had to be binarized from the ground-truth label distributions in the original datasets so as to implement LE algorithms and measure the similarity between the recovered label distributions and the ground-truths. To ensure the consistency of evaluation, we binarized the logical labels through the way in \cite{xu2018label} in this section.
	
	\subsection{Experimental Settings}
	To fully investigate the performance of our algorithms, i.e., LESC and gLESC, five state-of-the-art algorithms, including, FCM \cite{el2006study}, KM \cite{KM}, LP \cite{li2015leveraging}, ML \cite{hou2016multi}, and GLLE \cite{xu2019label} are employed. We list the parameter settings here. The parameters ${\lambda}_{1} $ and ${\lambda}_{2}$ are selected among $\{0.0001,0.001,...,10\}$ in our LESC and gLESC. In consistent with the parameters used in \cite{xu2019label}, the parameter $\alpha $ in LP is fixed to be 0.5, Gaussian kernel is employed in KM, the number of neighbors $K$  for ML is assigned to be $o+1$, and the parameter $\beta $ in FCM is fixed to be 2. Regarding to GLLE, the number of neighbors $K$  is assigned to be $o+1$ and the optimal value of parameter $\lambda $ is chosen from $\{0.01,0.1,...,100\}$. 
	
	\begin{table*}[!h]
		\centering
		\caption{Recovery Results With a Ten-fold Cross Validation (mean$\pm$std(rank))}
		\label{CVCV}
		\begin{center}
			\resizebox{0.85\textwidth}!{
				\begin{tabular}{c|c|c|c||c|c|c}
					\toprule
					\multirow{2}{*}{Dataset} &
					\multicolumn{3}{c||}{Measure Results by Cheb $\downarrow $}&
					\multicolumn{3}{c}{Measure Results by Canber $\downarrow $}\\
					\cmidrule{2-7}
					& GLLE & LESC & gLESC & GLLE & LESC & gLESC\\
					\midrule
					\tabincell{c}{Movie\\SBU\_3DFE\\SJAFFE\\Yeast-alpha\\Yeast-cdc\\Yeast-cold\\Yeast-diau\\Yeast-dtt\\Yeast-elu\\Yeast-heat\\Yeast-spo\\Yeast-spo5\\Yeast-spoem\\} &  		
					\tabincell{c}{0.1225$\pm$0.0026(3)\\0.1370$\pm$0.0051(3)\\0.1057$\pm$0.0111(3)\\0.0188$\pm$0.0029(3)\\0.0225$\pm$0.0005(3)\\0.0652$\pm$0.0021(3)\\0.0529$\pm$0.0013(3)\\0.0528$\pm$0.0016(3)\\0.0230$\pm$0.0004(3)\\0.0485$\pm$0.0014(3)\\0.0622$\pm$0.0027(3)\\0.0994$\pm$0.0041(3)\\0.0888$\pm$0.0045(3)\\}&
					\tabincell{c}{0.1205$\pm$0.0015(2)\\\bf{0.1218$\pm$0.0050(1)}\\0.0669$\pm$0.0057(2)\\0.0156$\pm$0.0004(2)\\0.0182$\pm$0.0007(2)\\0.0548$\pm$0.0023(2)\\0.0419$\pm$0.0011(2)\\0.0457$\pm$0.0019(2)\\0.0188$\pm$0.0003(2)\\0.0456$\pm$0.0008(2)\\0.0603$\pm$0.0022(2)\\0.0927$\pm$0.0036(2)\\0.0860$\pm$0.0040(2)\\}& \tabincell{c}{\bf{0.1199$\pm$0.0014(1)}\\0.1241$\pm$0.0048(2)\\\bf{0.0667$\pm$0.0091(1)}\\\bf{0.0144$\pm$0.0004(1)}\\\bf{0.0177$\pm$0.0005(1)}\\\bf{0.0527$\pm$0.0019(1)}\\\bf{0.0400$\pm$0.0010(1)}\\\bf{0.0379$\pm$0.0017(1)}\\\bf{0.0170$\pm$0.0006(1)}\\\bf{0.0437$\pm$0.0009(1)}\\\bf{0.0602$\pm$0.0016(1)}\\\bf{0.0908$\pm$0.0028(1)}\\\bf{0.0840$\pm$0.0038(1)}\\}&
					\tabincell{c}{1.0537$\pm$0.0208(3)\\0.8700$\pm$0.0254(3)\\0.7573$\pm$0.0625(3)\\1.1982$\pm$0.2159(3)\\0.9647$\pm$0.0164(3)\\0.3030$\pm$0.0106(3)\\0.6707$\pm$0.0132(3)\\0.2507$\pm$0.0084(3)\\0.8930$\pm$0.0107(3)\\0.4253$\pm$0.0099(3)\\0.5562$\pm$0.0217(3)\\0.3056$\pm$0.0136(3)\\0.1831$\pm$0.0097(3)\\}& 
					\tabincell{c}{1.0345$\pm$0.0145(2)\\\bf{0.8043$\pm$0.0222(1)}\\0.5542$\pm$0.0374(2)\\0.8509$\pm$0.0159(2)\\0.7450$\pm$0.0198(2)\\0.2578$\pm$0.0107(2)\\0.5039$\pm$0.0109(2)\\0.2177$\pm$0.0108(2)\\0.7188$\pm$0.0116(2)\\0.3995$\pm$0.0064(2)\\0.5312$\pm$0.0208(2)\\0.2867$\pm$0.0112(2)\\0.1782$\pm$0.0079(2)\\}&
					\tabincell{c}{\bf{1.0312$\pm$0.0143(1)}\\0.8277$\pm$0.0259(2)\\\bf{0.5324$\pm$0.0381(1)}\\\bf{0.7619$\pm$0.0154(1)}\\\bf{0.7065$\pm$0.0128(1)}\\\bf{0.2477$\pm$0.0098(1)}\\\bf{0.4584$\pm$0.0137(1)}\\\bf{0.1764$\pm$0.0077(1)}\\\bf{0.6254$\pm$0.0152(1)}\\\bf{0.3773$\pm$0.0094(1)}\\\bf{0.5254$\pm$0.0098(1)}\\\bf{0.2796$\pm$0.0094(1)}\\\bf{0.1738$\pm$0.0080(1)}\\}
					\\
					\midrule
					Avg.Rank & 3.00 & 1.92 & \bf{1.08} & 3.00 & 1.92 & \bf{1.08} \\
					\midrule
					\multirow{2}{*}{Dataset} &
					\multicolumn{3}{c||}{Measure Results by Clark $\downarrow $}&
					\multicolumn{3}{c}{Measure Results by KL $\downarrow $}\\
					\cmidrule{2-7}
					& GLLE & LESC & gLESC & GLLE & LESC & gLESC\\
					\midrule
					\tabincell{c}{Movie\\SBU\_3DFE\\SJAFFE\\Yeast-alpha\\Yeast-cdc\\Yeast-cold\\Yeast-diau\\Yeast-dtt\\Yeast-elu\\Yeast-heat\\Yeast-spo\\Yeast-spo5\\Yeast-spoem\\} &  	
					\tabincell{c}{0.5721$\pm$0.0106(3)\\0.4038$\pm$0.0113(3)\\0.3697$\pm$0.0253(3)\\0.3491$\pm$0.0560(3)\\0.3092$\pm$0.0065(3)\\0.1742$\pm$0.0062(3)\\0.2954$\pm$0.0053(3)\\0.1446$\pm$0.0047(3)\\0.2922$\pm$0.0038(3)\\0.2111$\pm$0.0048(3)\\0.2685$\pm$0.0107(3)\\0.1962$\pm$0.0093(3)\\0.1313$\pm$0.0071(3)\\}& \tabincell{c}{0.5640$\pm$0.0074(2)\\\bf{0.3792$\pm$0.0100(1)}\\0.2722$\pm$0.0208(2)\\0.2553$\pm$0.0053(2)\\0.2457$\pm$0.0068(2)\\0.1491$\pm$0.0066(2)\\0.2318$\pm$0.0053(2)\\0.1255$\pm$0.0058(2)\\0.2396$\pm$0.0044(2)\\0.1987$\pm$0.0030(2)\\0.2577$\pm$0.0090(2)\\0.1865$\pm$0.0073(2)\\0.1281$\pm$0.0056(2)\\}&
					\tabincell{c}{\bf{0.5621$\pm$0.0071(1)}\\0.3895$\pm$0.0114(2)\\\bf{0.2628$\pm$0.0178(1)}\\\bf{0.2313$\pm$0.0054(1)}\\\bf{0.2351$\pm$0.0056(1)}\\\bf{0.1441$\pm$0.0057(1)}\\\bf{0.2149$\pm$0.0066(1)}\\\bf{0.1025$\pm$0.0046(1)}\\\bf{0.2110$\pm$0.0059(1)}\\\bf{0.1892$\pm$0.0041(1)}\\\bf{0.2548$\pm$0.0049(1)}\\\bf{0.1809$\pm$0.0065(1)}\\\bf{0.1248$\pm$0.0058(1)}\\}& \tabincell{c}{0.1248$\pm$0.0054(3)\\0.0797$\pm$0.0042(3)\\0.0531$\pm$0.0079(3)\\0.0142$\pm$0.0045(3)\\0.0141$\pm$0.0006(3)\\0.0181$\pm$0.0013(3)\\0.0269$\pm$0.0010(3)\\0.0128$\pm$0.0009(3)\\0.0132$\pm$0.0004(3)\\0.0169$\pm$0.0008(3)\\0.0293$\pm$0.0026(3)\\0.0338$\pm$0.0032(3)\\0.0268$\pm$0.0027(3)\\}&
					\tabincell{c}{0.1198$\pm$0.0027(2)\\\bf{0.0648$\pm$0.0037(1)}\\0.0271$\pm$0.0039(2)\\0.0080$\pm$0.0004(2)\\0.0092$\pm$0.0006(2)\\0.0140$\pm$0.0012(2)\\0.0176$\pm$0.0008(2)\\0.0100$\pm$0.0011(2)\\0.0091$\pm$0.0004(2)\\0.0152$\pm$0.0006(2)\\0.0275$\pm$0.0020(2)\\0.0318$\pm$0.0023(2)\\0.0265$\pm$0.0023(2)\\}&
					\tabincell{c}{\bf{0.1189$\pm$0.0023(1)}\\0.0677$\pm$0.0042(2)\\\bf{0.0266$\pm$0.0052(1)}\\\bf{0.0066$\pm$0.0003(1)}\\\bf{0.0086$\pm$0.0005(1)}\\\bf{0.0136$\pm$0.0014(1)}\\\bf{0.0156$\pm$0.0010(1)}\\\bf{0.0068$\pm$0.0007(1)}\\\bf{0.0070$\pm$0.0004(1)}\\\bf{0.0138$\pm$0.0005(1)}\\\bf{0.0270$\pm$0.0008(1)}\\\bf{0.0301$\pm$0.0023(1)}\\\bf{0.0254$\pm$0.0021(1)}\\}
					\\
					\midrule
					Avg.Rank & 3.00 & 1.92 & \bf{1.08} & 3.00 & 1.92 & \bf{1.08} \\
					\midrule
					\multirow{2}{*}{Dataset} &
					\multicolumn{3}{c||}{Measure Results by Cosine $\uparrow $}&
					\multicolumn{3}{c}{Measure Results by Intersec $\uparrow $}\\
					\cmidrule{2-7}
					& GLLE & LESC & gLESC & GLLE & LESC & gLESC\\
					\midrule
					\tabincell{c}{Movie\\SBU\_3DFE\\SJAFFE\\Yeast-alpha\\Yeast-cdc\\Yeast-cold\\Yeast-diau\\Yeast-dtt\\Yeast-elu\\Yeast-heat\\Yeast-spo\\Yeast-spo5\\Yeast-spoem\\} &
					\tabincell{c}{0.9350$\pm$0.0026(3)\\0.9138$\pm$0.0047(3)\\0.9466$\pm$0.0091(3)\\0.9862$\pm$0.0045(3)\\0.9868$\pm$0.0004(3)\\0.9827$\pm$0.0013(3)\\0.9750$\pm$0.0010(3)\\0.9879$\pm$0.0008(3)\\0.9872$\pm$0.0003(3)\\0.9842$\pm$0.0008(3)\\0.9735$\pm$0.0022(3)\\0.9706$\pm$0.0022(3)\\0.9780$\pm$0.0019(3)\\}&
					\tabincell{c}{0.9375$\pm$0.0018(2)\\\bf{0.9315$\pm$0.0042(1)}\\0.9745$\pm$0.0036(2)\\0.9922$\pm$0.0003(2)\\0.9914$\pm$0.0006(2)\\0.9869$\pm$0.0012(2)\\0.9843$\pm$0.0006(2)\\0.9905$\pm$0.0009(2)\\0.9913$\pm$0.0003(2)\\0.9858$\pm$0.0005(2)\\0.9751$\pm$0.0017(2)\\0.9734$\pm$0.0018(2)\\0.9784$\pm$0.0017(2)\\}& \tabincell{c}{\bf{0.9380$\pm$0.0017(1)}\\0.9287$\pm$0.0044(2)\\\bf{0.9752$\pm$0.0055(1)}\\\bf{0.9936$\pm$0.0003(1)}\\\bf{0.9921$\pm$0.0003(1)}\\\bf{0.9878$\pm$0.0010(1)}\\\bf{0.9865$\pm$0.0008(1)}\\\bf{0.9936$\pm$0.0006(1)}\\\bf{0.9933$\pm$0.0004(1)}\\\bf{0.9871$\pm$0.0004(1)}\\\bf{0.9754$\pm$0.0009(1)}\\\bf{0.9745$\pm$0.0015(1)}\\\bf{0.9796$\pm$0.0015(1)}\\}& 
					\tabincell{c}{0.8290$\pm$0.0033(3)\\0.8426$\pm$0.0050(3)\\0.8715$\pm$0.0119(3)\\0.9341$\pm$0.0123(3)\\0.9362$\pm$0.0009(3)\\0.9246$\pm$0.0025(3)\\0.9056$\pm$0.0021(3)\\0.9381$\pm$0.0021(3)\\0.9364$\pm$0.0008(3)\\0.9303$\pm$0.0017(3)\\0.9084$\pm$0.0036(3)\\0.9006$\pm$0.0041(3)\\0.9112$\pm$0.0045(3)\\}& 
					\tabincell{c}{0.8330$\pm$0.0023(2)\\\bf{0.8544$\pm$0.0045(1)}\\0.9070$\pm$0.0065(2)\\0.9529$\pm$0.0009(2)\\0.9509$\pm$0.0014(2)\\0.9363$\pm$0.0025(2)\\0.9300$\pm$0.0014(2)\\0.9463$\pm$0.0027(2)\\0.9491$\pm$0.0008(2)\\0.9342$\pm$0.0011(2)\\0.9124$\pm$0.0034(2)\\0.9073$\pm$0.0036(2)\\0.9140$\pm$0.0040(2)\\}&
					\tabincell{c}{\bf{0.8338$\pm$0.0023(1)}\\0.8501$\pm$0.0050(2)\\\bf{0.9105$\pm$0.0081(1)}\\\bf{0.9579$\pm$0.0008(1)}\\\bf{0.9535$\pm$0.0008(1)}\\\bf{0.9390$\pm$0.0023(1)}\\\bf{0.9365$\pm$0.0018(1)}\\\bf{0.9563$\pm$0.0019(1)}\\\bf{0.9558$\pm$0.0011(1)}\\\bf{0.9380$\pm$0.0015(1)}\\\bf{0.9134$\pm$0.0017(1)}\\\bf{0.9092$\pm$0.0028(1)}\\\bf{0.9160$\pm$0.0038(1)}\\}
					\\
					\midrule
					Avg.Rank & 3.00 & 1.92 & \bf{1.08} & 3.00 & 1.92 & \bf{1.08} \\
					\bottomrule
				\end{tabular}
			}
		\end{center}
	\end{table*}
	
	Since both the recovered and ground-truth label distributions are label vectors, the average distance or similarity between them is calculated to evaluate the LE algorithms thoroughly. For a fair comparison, six measures are selected, where the first four are distance-based measures and the last two are similarity-based measures, reflecting the performance of LE algorithms from different aspects in semantics. As shown in Table \ref{Evaluation_Measures} where $\hat{D}$ denotes the real label distributions, for these metrics, i.e., Chebyshev distance (Cheb), Canberra metric (Canber), Clark distance (Clark), Kullback-Leibler divergence (KL), cosine coefficient (Cosine), and intersection similarity (Intersec), $\downarrow $ states "the smaller the greater", and $\uparrow $ states "the larger the greater".
	
	\subsection{Experimental Results}
	Firstly, we present the recovery performance on the artificial dataset, and to illustrate the recovery performance visually, the three-dimensional label distributions are separately converted into the RGB color channels, which are reinforced by the decorrelation stretch process for easier observation. In other words, the label distribution of each point in the feature space can be represented by its color. As shown in Fig.~\ref{Art_3D}, the color patterns can be directly observed to compare both the ground truth and the recovered label distributions of different methods. 
	
	To further investigate the recovery performance, we present the quantitative results of these aforementioned algorithms in metrics of Cheb, Canber, Clark, KL, Cosine, and Intersec (as shown in Table~\ref{Recovery_Results_Cheb_Canber}). 
	
	To show that the improvements of our methods are indeed statistically significant, experiments, which adopt the ten-fold cross validation and T-test, are also conducted in this section. It is notable that the cross validation is not suitable for FCM, KM, LP, and ML, since they directly recover label distributions based on all instances rather than learn the specific recovery model or mapping function. Therefore, for the sake of fairness, we run experiments with a ten-fold cross validation on GLLE, LESC, and gLESC for comparison. The experimental results are reported in Table~\ref{CVCV} and~\ref{T_testT_test}. 
	
	\section{Analysis and Discussion}
	In this section, we focus on the analysis and discussion of experimental results. 
	\begin{table*}[!h]
		\centering
		\caption{T-test Results}
		\label{T_testT_test}
		\begin{center}
			\resizebox{0.92\textwidth}!{
				\begin{tabular}{c|c|c|c|c|c|c||c|c|c|c|c|c}
					\toprule
					\multirow{3}{*}{Dataset} &
					\multicolumn{6}{c||}{Measure Results by Cheb}&
					\multicolumn{6}{c}{Measure Results by Canber}\\
					\cmidrule{2-13}
					& \multicolumn{2}{c|}{GLLE and LESC} & \multicolumn{2}{c|}{GLLE and gLESC} & \multicolumn{2}{c||}{LESC and gLESC} & \multicolumn{2}{c|}{GLLE and LESC} & \multicolumn{2}{c|}{GLLE and gLESC} & \multicolumn{2}{c}{LESC and gLESC}\\
					\cmidrule{2-13}
					& \textit{t}-value & \textit{p} & \textit{t}-value & \textit{p} & \textit{t}-value & \textit{p} & \textit{t}-value & \textit{p} & \textit{t}-value & \textit{p} & \textit{t}-value & \textit{p}\\
					\midrule
					\tabincell{c}{Movie\\SBU\_3DFE\\SJAFFE\\Yeast-alpha\\Yeast-cdc\\Yeast-cold\\Yeast-diau\\Yeast-dtt\\Yeast-elu\\Yeast-heat\\Yeast-spo\\Yeast-spo5\\Yeast-spoem\\} &  		
					\tabincell{c}{2.8896\\6.0865\\12.9504\\3.3151\\12.803\\18.0582\\18.6861\\9.6781\\23.7774\\5.1934\\1.5796\\4.4747\\1.5213\\}&
					\tabincell{c}{0.0179\\1.82E-04\\4.01E-07\\0.009\\4.43E-07\\2.23E-08\\1.65E-08\\4.70E-06\\1.96E-09\\5.69E-04\\0.1487\\0.0015\\0.1625\\}& \tabincell{c}{3.7356\\7.5646\\8.9985\\4.5912\\46.4826\\11.3127\\25.0726\\24.3695\\28.7222\\9.7099\\2.0551\\5.1503\\2.2097\\}&
					\tabincell{c}{0.0047\\3.45E-05\\8.55E-06\\0.0013\\4.94E-12\\1.27E-06\\1.23E-09\\1.58E-09\\3.66E-10\\4.57E-06\\0.07\\6.03E-04\\0.0545\\}& 
					\tabincell{c}{10.1806\\-0.906\\70.0691\\5.3743\\1.63\\1.7457\\4.9986\\16.6552\\8.4796\\5.5428\\0.132\\2.0249\\0.9828\\}&
					\tabincell{c}{3.08E-06\\0.3882\\0.9464\\4.48E-04\\0.1375\\0.1148\\7.40E-04\\4.53E-08\\1.39E-05\\3.60E-04\\0.8979\\0.0735\\0.3514\\}&
					\tabincell{c}{4.1985\\6.0392\\16.8084\\4.9394\\21.9595\\27.6061\\30.2152\\7.6114\\32.2289\\6.5476\\2.458\\3.9804\\1.2646\\}&
					\tabincell{c}{0.0023\\1.93E-04\\4.18E-08\\8.03E-04\\3.98E-09\\5.21E-10\\2.33E-10\\3.29E-05\\1.31E-10\\1.05E-04\\0.0363\\0.0032\\0.2378\\}&
					\tabincell{c}{4.9693\\4.8465\\10.0808\\6.1692\\91.6064\\9.0292\\35.6724\\24.5951\\54.6123\\10.8416\\3.6905\\4.8572\\1.9638\\}&
					\tabincell{c}{7.71E-04\\9.13E-04\\3.35E-06\\1.65E-04\\1.12E-14\\8.31E-06\\5.29E-11\\1.46E-09\\1.16E-12\\1.82E-06\\0.005\\8.99E-04\\0.0811\\}&
					\tabincell{c}{14.1472\\-2.1394\\1.6848\\12.2863\\4.2433\\1.6152\\8.5495\\15.0992\\16.6537\\10.7182\\0.6831\\2.1382\\1.0861\\}&
					\tabincell{c}{1.87E-07\\0.061\\10.1263\\6.30E-07\\0.0022\\0.1407\\1.30E-05\\1.07E-07\\4.53E-08\\2.00E-06\\0.5117\\0.0612\\0.3057\\}
					\\
					\midrule
					\multirow{3}{*}{Dataset} &
					\multicolumn{6}{c||}{Measure Results by Clark}&
					\multicolumn{6}{c}{Measure Results by KL}\\
					\cmidrule{2-13}
					& \multicolumn{2}{c|}{GLLE and LESC} & \multicolumn{2}{c|}{GLLE and gLESC} & \multicolumn{2}{c||}{LESC and gLESC} & \multicolumn{2}{c|}{GLLE and LESC} & \multicolumn{2}{c|}{GLLE and gLESC} & \multicolumn{2}{c}{LESC and gLESC}\\
					\cmidrule{2-13}
					& \textit{t}-value & \textit{p} & \textit{t}-value & \textit{p} & \textit{t}-value & \textit{p} & \textit{t}-value & \textit{p} & \textit{t}-value & \textit{p} & \textit{t}-value & \textit{p} \\
					\midrule
					\tabincell{c}{Movie\\SBU\_3DFE\\SJAFFE\\Yeast-alpha\\Yeast-cdc\\Yeast-cold\\Yeast-diau\\Yeast-dtt\\Yeast-elu\\Yeast-heat\\Yeast-spo\\Yeast-spo5\\Yeast-spoem\\} &  		
					\tabincell{c}{3.3634\\5.0973\\21.104\\5.0905\\17.2728\\21.8715\\25.6799\\7.9846\\28.5291\\6.3421\\2.2812\\3.0832\\1.149\\}&
					\tabincell{c}{0.0083\\6.48E-04\\5.65E-09\\6.53E-04\\3.29E-08\\4.12E-09\\9.92E-10\\2.25E-05\\3.89E-10\\1.34E-04\\0.0485\\0.0131\\0.2802\\}&\tabincell{c}{4.0836\\3.6114\\11.0581\\6.3699\\120.2465\\8.4552\\30.1119\\23.0548\\42.0784\\10.8758\\3.2854\\4.3741\\1.8594\\}&
					\tabincell{c}{0.0027\\5.60E-03\\1.54E-06\\1.30E-04\\9.66E-16\\1.42E-05\\2.40E-10\\2.58E-09\\1.21E-11\\1.77E-06\\0.0094\\1.80E-03\\0.0959\\}& 
					\tabincell{c}{17.2405\\-2.1287\\1.5257\\9.2764\\3.0326\\1.3457\\6.7889\\15.2243\\12.2026\\8.2975\\0.7516\\2.4055\\1.1265\\}&
					\tabincell{c}{3.35E-08\\0.0622\\0.1614\\6.66E-06\\0.0142\\0.2113\\8.01E-05\\9.92E-08\\6.68E-07\\1.65E-05\\0.4715\\0.0395\\0.2891\\}&
					\tabincell{c}{3.5518\\7.7975\\13.8502\\4.212\\15.1542\\17.5623\\21.9401\\5.9598\\22.3721\\5.9554\\1.6755\\1.7095\\0.3615\\}&
					\tabincell{c}{0.0062\\2.72E-05\\2.25E-07\\2.30E-03\\1.03E-07\\2.85E-08\\4.01E-09\\2.13E-04\\3.37E-09\\2.14E-04\\0.1282\\0.1215\\0.726\\}&
					\tabincell{c}{4.1636\\8.1073\\8.8168\\5.1794\\82.0411\\5.8644\\27.0149\\18.4522\\43.9034\\9.7614\\2.4652\\2.9055\\1.1862\\}&
					\tabincell{c}{0.0024\\1.99E-05\\1.01E-05\\5.80E-04\\3.01E-14\\2.39E-04\\6.32E-10\\1.85E-08\\8.24E-12\\4.37E-06\\0.0359\\1.74E-02\\0.2659\\}&
					\tabincell{c}{13.0287\\-1.5343\\0.3674\\8.4349\\1.8957\\0.5674\\5.1367\\8.9697\\11.369\\6.6008\\0.6443\\1.8091\\1.1077\\}&
					\tabincell{c}{3.81E-07\\0.1593\\0.7218\\1.45E-05\\0.0905\\0.5843\\6.14E-04\\8.78E-06\\1.22E-06\\9.92E-05\\0.5355\\0.1039\\0.2967\\}
					\\
					\midrule
					\multirow{3}{*}{Dataset} &
					\multicolumn{6}{c||}{Measure Results by Cosine }&
					\multicolumn{6}{c}{Measure Results by Intersec }\\
					\cmidrule{2-13}
					& \multicolumn{2}{c|}{GLLE and LESC} & \multicolumn{2}{c|}{GLLE and gLESC} & \multicolumn{2}{c||}{LESC and gLESC} & \multicolumn{2}{c|}{GLLE and LESC} & \multicolumn{2}{c|}{GLLE and gLESC} & \multicolumn{2}{c}{LESC and gLESC}\\
					\cmidrule{2-13}
					& \textit{t}-value & \textit{p} & \textit{t}-value & \textit{p} & \textit{t}-value & \textit{p} & \textit{t}-value & \textit{p} & \textit{t}-value & \textit{p} & \textit{t}-value & \textit{p} \\
					\midrule
					\tabincell{c}{Movie\\SBU\_3DFE\\SJAFFE\\Yeast-alpha\\Yeast-cdc\\Yeast-cold\\Yeast-diau\\Yeast-dtt\\Yeast-elu\\Yeast-heat\\Yeast-spo\\Yeast-spo5\\Yeast-spoem\\} &  		
					\tabincell{c}{3.7784\\8.33\\11.4564\\15.5619\\18.2035\\18.6097\\24.3265\\7.0969\\24.6254\\6.401\\1.6114\\3.5837\\0.5809\\}&
					\tabincell{c}{0.0044\\1.60E-05\\1.14E-06\\8.19E-08\\2.08E-08\\1.71E-08\\1.60E-09\\5.69E-05\\1.44E-09\\1.25E-04\\1.42E-01\\0.0059\\0.5755\\}& \tabincell{c}{4.5154\\9.1258\\8.6359\\5.0784\\86.5044\\7.5981\\31.0805\\20.906\\54.8897\\10.6537\\2.4075\\4.304\\1.8677\\}&
					\tabincell{c}{0.0015\\7.62E-06\\1.20E-05\\6.64E-04\\1.87E-14\\3.33E-05\\1.81E-10\\6.14E-09\\1.11E-12\\2.11E-06\\0.0394\\2.00E-03\\0.0946\\}& 
					\tabincell{c}{9.704\\-1.3034\\0.4266\\8.5692\\2.8618\\1.3547\\7.1992\\11.6758\\13.1411\\10.7267\\0.4233\\2.0049\\1.483\\}&
					\tabincell{c}{4.59E-06\\0.2248\\0.6797\\1.27E-05\\0.0187\\0.2085\\5.09E-05\\9.72E-07\\3.54E-07\\1.99E-06\\0.682\\0.0759\\0.1722\\}&
					\tabincell{c}{4.9195\\5.2726\\13.4534\\4.6951\\23.447\\28.1308\\30.2034\\7.7614\\31.7878\\6.1048\\2.3603\\4.4747\\1.5213\\}&
					\tabincell{c}{8.25E-04\\5.12E-04\\2.89E-07\\1.10E-03\\2.22E-09\\4.41E-10\\2.34E-10\\2.82E-05\\1.48E-10\\1.78E-04\\0.0426\\0.0015\\0.1625\\}&
					\tabincell{c}{5.945\\4.2478\\8.9585\\5.9112\\101.2241\\10.1672\\37.6592\\25.134\\57.1208\\10.2376\\3.6816\\5.1503\\2.2097\\}&
					\tabincell{c}{2.17E-04\\2.10E-03\\8.87E-06\\2.26E-04\\4.55E-15\\3.12E-06\\3.26E-11\\1.20E-09\\7.78E-13\\2.94E-06\\0.0051\\6.03E-04\\0.0545\\}&
					\tabincell{c}{12.6741\\-1.8545\\1.2622\\12.2582\\4.2933\\1.8507\\9.2532\\15.2458\\17.7899\\10.4932\\0.6842\\2.0249\\0.9828\\}&
					\tabincell{c}{4.83E-07\\0.0967\\0.2386\\6.42E-07\\0.002\\0.0972\\6.80E-06\\9.80E-08\\2.54E-08\\2.39E-06\\0.5111\\0.0735\\0.3514\\}
					\\
					\bottomrule
				\end{tabular}
			}
		\end{center}
	\end{table*}	
	
	\subsection{Analysis of Experimental Results}
\begin{figure*}[!t]
		\centering	
		\resizebox{1\textwidth}!{
			\includegraphics{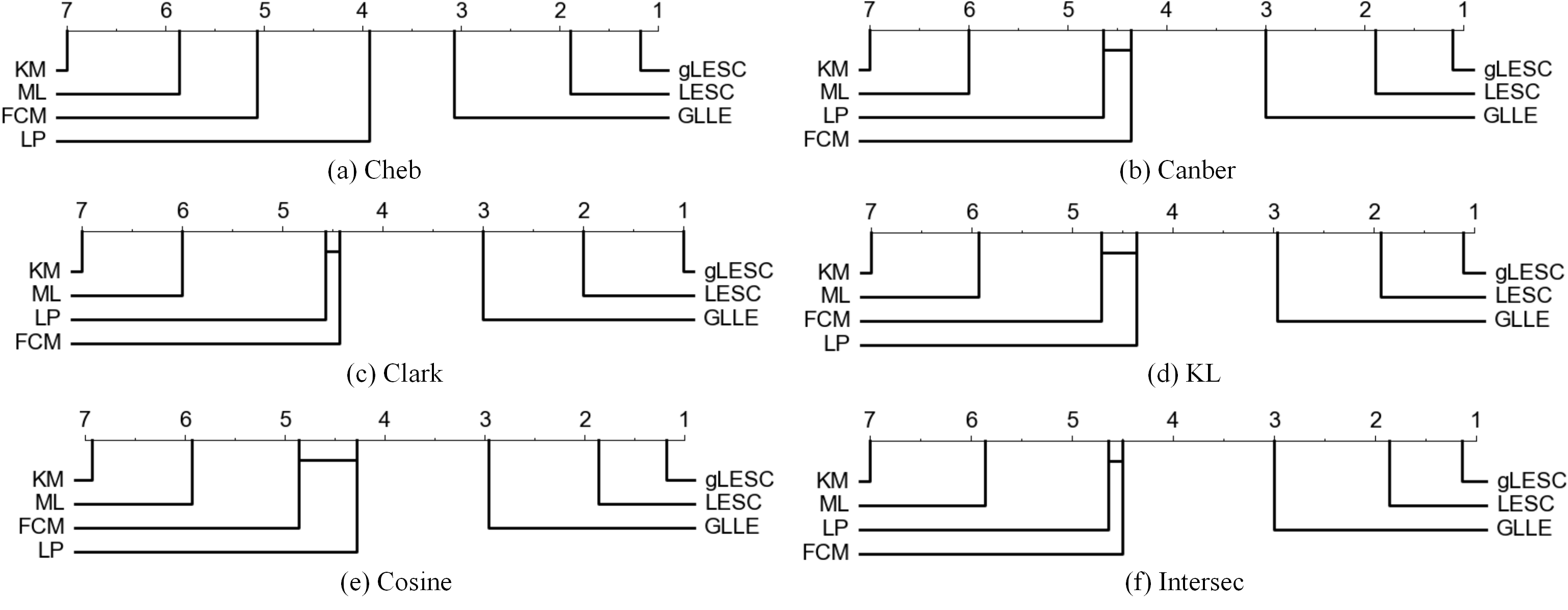}
		}
		\caption{CD diagrams of different LE methods on six measures, including Cheb, Canber, Clark, KL, Cosine, and Intersec. CD diagrams are calculated based on the Wilcoxon-Holm method \cite{CD_diagram}. Specifically, the method located on the right side is better that the method on the left side, and the line between two methods denotes that their recovery results are different within one critical difference.}
		\label{CD_diagram}
	\end{figure*}

	As illustrated by the Table~\ref{Recovery_Results_Cheb_Canber}, the recovery performance can be ranked as gLESC$>$LESC$>$GLLE$>$LP$>$FCM$>$ML$>$KM in general.  Additionally, the critical difference (CD) of average rank is also reported in Fig.~\ref{CD_diagram}, it can be observed that gLESC achieves the optimal recovery results on all metrics, and the proposed LESC also attain the sub-optimal performance.
	
	\subsubsection{Comparison with FCM and KM}
	The recovery results of FCM and KM are shown in Fig.~\ref{Art_3D},~\ref{CD_diagram}, and Table~\ref{Recovery_Results_Cheb_Canber}. We can see that,  compared with FCM and KM, the proposed LESC and gLESC perform much better in all the metrics of evaluation measures. Taking experimental results on Table~\ref{Recovery_Results_Cheb_Canber} for example, $0.087$ and $0.084$ are achieved by LESC and gLESC, while $0.233$ and $0.408$ are obtained by FCM and KM on Yeast-spoem dataset in the metric of Cheb. The average rank of LESC and gLESC is $2.00$ and $1.00$, comparing to $4.43$ and $7.00$ of FCM and KM in the metric of Clark. Actually, as introduced in Section~\ref{Section2}, although the topological structures of the feature space and membership degrees of different labels are employed in FCM and KM to recover label distributions, they lack of a good investigation of the sample correlations of samples for the recovery process. 
	
	\subsubsection{Comparison with LP, ML, and GLLE}
	In a big picture, LESC, and gLESC achieve more promising recovery performance on all datasets, which can be observed in Table~\ref{Recovery_Results_Cheb_Canber} and Fig.~\ref{CD_diagram} obviously. The main difference between the proposed methods and these approaches is that the global topological structure is investigated in the proposed LESC and gLESC, while the local structure is used in LP, ML and GLLE. Here we emphatically analysis experimental results of GLLE, LESC and gLESC, since recovery results of GLLE are better than LP and ML on the whole. Here are some statistics. On the SJAFFE dataset, compared to the results achieved by GLLE, the results of LESC indicate an increase of $0.015$ and $0.033$, and the results og gLESC attain an increase of $0.017$ and $0.037$ in metrics of Consine and Intersec. 
	
	Furthermore, we also discuss results of experiments with a ten-fold cross validation and T-test are discussed to show that the improvements achieved by LESC and gLESC are indeed statistically significant. As shown in Table~\ref{CVCV}, we can see that LESC and gLESC outperform GLLE as well. The reason is that the intrinsic global sample correlations can be employed in LESC and gLESC for the recovery of label distribution, while local graph information is used in GLLE. Results of T-test are also provided in Table~\ref{T_testT_test}, and we set significance level to 0.05. In comparison between GLLE and LESC, statistically significant differences can be obtained in most cases. For example, on the Movie dataset, statistically significant improvements can be observed clearly on all evaluation measures (Cheb: p=0.0179; Canber: p=0.0023; Clark: p=0.0083; KL: p=0.0062; Cosine: p=0.0044; Intersec: p=0.0008). A similar observation can be achieved in comparison between GLLE and gLESC. 
	
	{\bf{Are sample correlations obtained by the low-rank representation suitable for LE?}} As analyzed and discussed before, we can conclude that LESC and gLESC, which leverage the low-rank representation to attain global sample correlations for LE, outperform GLLE, which use the distance-based similarity to achieve label recovery, by a large margin. Consequently, the sample correlations obtained by the low-rank representation are suitable for LE.
	
	\subsubsection{Comparison between LESC and gLESC}
	In general, the performance of gLESC is better than LESC, which can be observed from Table~\ref{Recovery_Results_Cheb_Canber},~\ref{CVCV}, and~\ref{T_testT_test} obviously. For most cases, gLESC is more general and better than LESC. The reason is that the global sample correlations in both the feature space and label space are explored in gLESC, while only information in feature space is utilized in LESC.  
	
	{\bf{Are sample correlations captured from both the feature space and label space better for LE?}} Comparing to LESC, gLESC leverages a tensor multi-rank minimization to obtain the sample correlations from both the feature space and label space. Since the sample correlations used in gLSEC are more comprehensive than that in LESC, it is expected that gLESC can attain better recovery performance. From the quantitative experimental results, we can conclude that sample correlations captured from both the feature space and label space are better for LE. 
	
	\begin{figure*}[!t]
		\centering	
		\resizebox{1\textwidth}!{
			\includegraphics{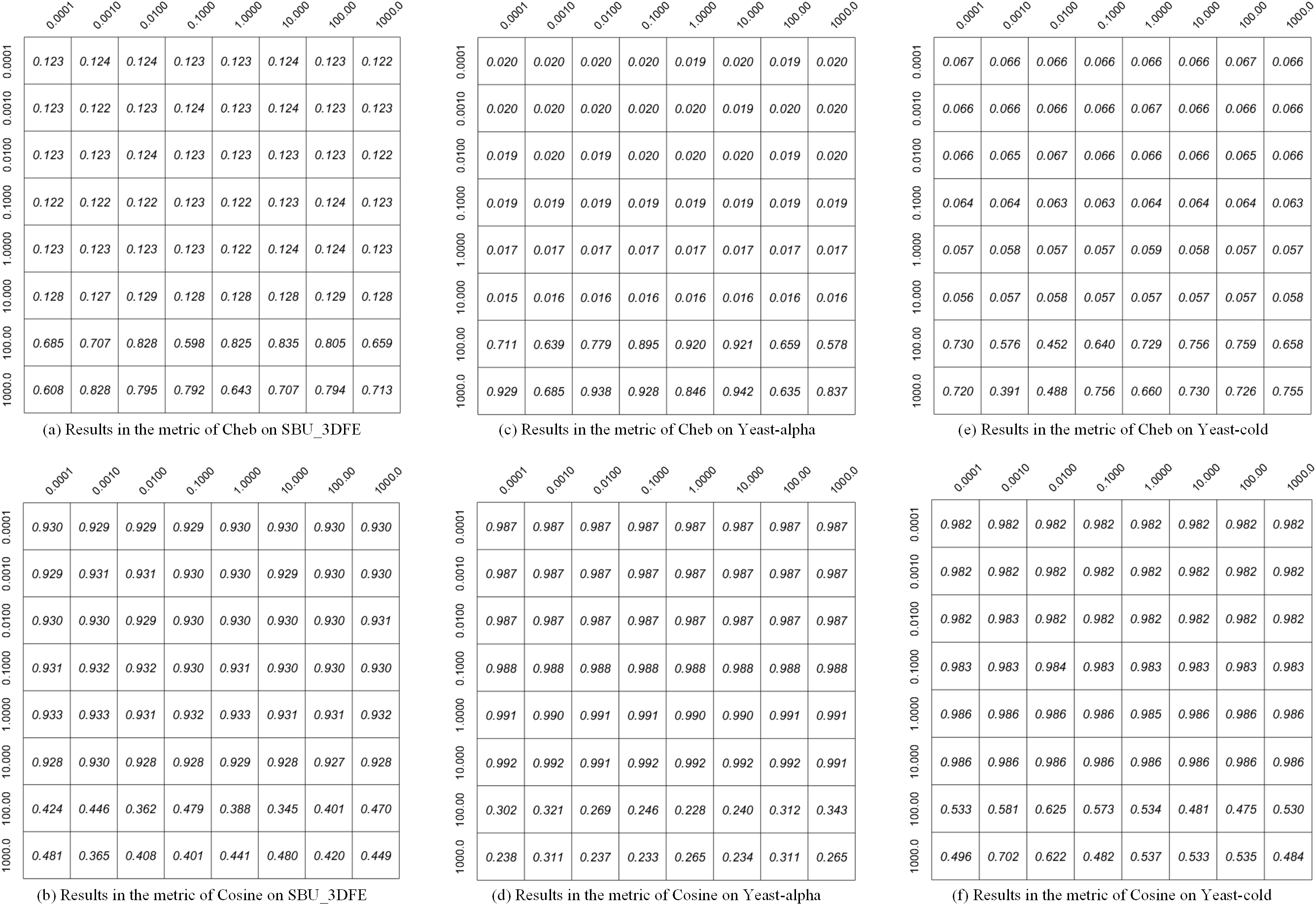}
		}
		\caption{Label recovery performance of LESC on SBU\_3DFE, Yeast-alpha, and Yeast-cold in metrics of Cheb and Cosine. Specifically, different rows denote different values of $\lambda_1$, and different columns denote different values of $\lambda_2$.}
		\label{ParameterSensitivity_LESC}
	\end{figure*}
	
	\begin{figure*}[!t]
		\centering	
		\resizebox{1\textwidth}!{
			\includegraphics{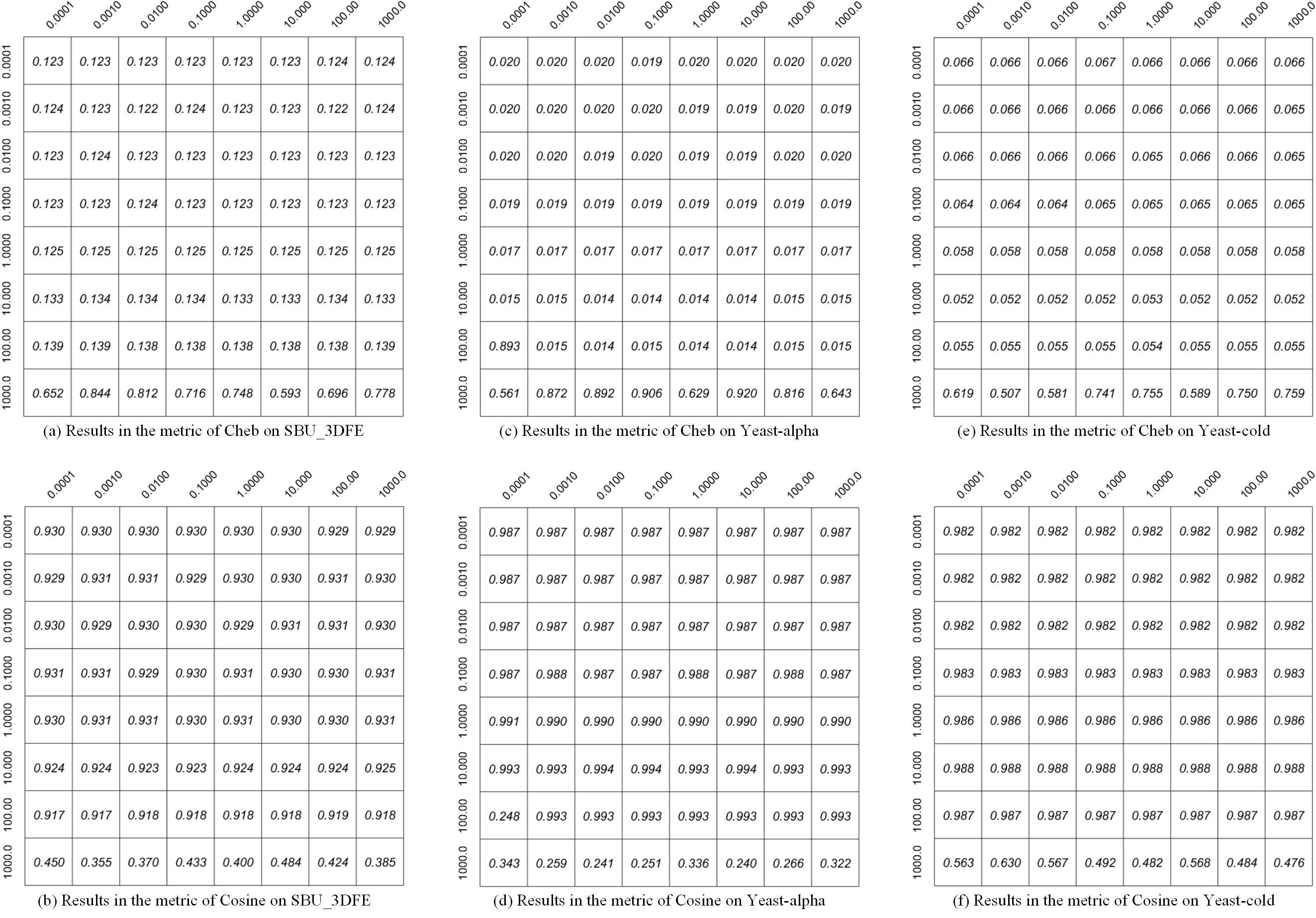}
		}
		\caption{Label recovery performance of gLESC on SBU\_3DFE, Yeast-alpha, and Yeast-cold in metrics of Cheb and Cosine. Specifically, different rows denote different values of $\lambda_1$, and different columns denote different values of $\lambda_2$.}
		\label{ParameterSensitivity_gLESC}
	\end{figure*}

	\subsection{Parameters Sensitivity}
	Two trade-off hyperparameters, including $\lambda_1$ and $\lambda_1$, are involved in our proposed methods. The influence of them is analyzed separately by fixing one parameter and tuning another one chosen from $\{0.0001,0.001,...,1000\}$. In this section, we take experimental results on SBU\_3DFE, Yeast-alpha, and Yeast-cold datasets in metrics of Cheb and Cosine for example, which can be seen in Fig.~\ref{ParameterSensitivity_LESC} and Fig.~\ref{ParameterSensitivity_gLESC}. Although only the cases of three datasets are illustrated here, the same observations can be obtained in other datasets.
	
	For LESC, when the low-rank coefficient ${\lambda}_{2}$ varies with the trade-off parameter ${\lambda}_{1} $ fixed, two shown measure results of the recovery performance fluctuates in a very tiny range that could not even be distinguished. As we increase the parameter ${\lambda}_{1} $ from 0.0001 to 0.1, the recovery performance also turns out to change within a small scope. When ${\lambda}_{1} $ is geared to 1 or 10, the results even zooms up to a higher level. Particularly, taking Yeast-alpha dataset for reference, it is found that when $\lambda_1$ is chosen from $\{0.0001,0.001,...,10\}$, our worst measure result still far exceeds that of the previous state-of-the-art baseline, i.e, 0.987 versus 0.973 (best result attained by GLLE) in the metric of Cosine. Regrading to gLESC, similar observations can be reached as well, and we skip them here for the compactness of this paper. As discussed before, these phenomena indicate that our algorithms, both LESC and gLESC, are robust when the values of ${\lambda}_{1}$ and ${\lambda}_{2}$ in the objective function vary by a large scope. This ensures us to generalize our algorithm to different datasets without much effort in terms of adjusting the values of hyperparameters in practice.
	
	\section{Conclusion}
	In this paper, two novel LE methods, i.e., LESC and gLESC, are proposed to boost the LE performance by exploiting the underlying sample correlations. LESC explores the low-rank representation from the feature space, and gLESC further investigates the sample correlations by utilizing a tensor multi-rank minimization to obtain more suitable sample correlations from both the feature space and label space during the label distribution recovery process. Extensive experimental results on 14 datasets show that LE can really benefit from the sample correlations. Experimental results demonstrate the remarkable superiority of the proposed LESC and gLESC over several state-of-the-art algorithms in recovering the label distributions. Further analysis on the influence of hyperparameters verifies the robustness of our methods. 
	
	\section*{Acknowledgment}
	This work is supported by the Fundamental Research Funds for Central Universities under Grant No. xzy012019045.
	
	
	%


	\ifCLASSOPTIONcaptionsoff
	\newpage
	\fi

	
	
	%
	
	\bibliographystyle{IEEEtran}
	\bibliography{gLESC_revised_manuscript}
	
\end{document}